\documentclass{article}

\usepackage{PRIMEarxiv}
\usepackage[numbers]{natbib}

\usepackage[utf8]{inputenc} % allow utf-8 input
\usepackage[T1]{fontenc}    % use 8-bit T1 fonts
\usepackage{hyperref}       % hyperlinks
\usepackage{url}            % simple URL typesetting
\usepackage{booktabs}       % professional-quality tables
\usepackage{amsfonts}       % blackboard math symbols
\usepackage{nicefrac}       % compact symbols for 1/2, etc.
\usepackage{microtype}      % microtypography
\usepackage{lipsum}
\usepackage{fancyhdr}       % header
\usepackage{graphicx}       % graphics
\graphicspath{{media/}}     % organize your images and other figures under media/ folder
\usepackage{caption}
\usepackage{setspace}
\usepackage{amssymb}
\usepackage{multirow}
\usepackage{lscape}
\usepackage{amsthm}
\usepackage{booktabs}       % professional-quality tables
\usepackage{amsmath}        % math environments
\usepackage{mathrsfs}
\usepackage{longtable}
\usepackage{url}
\usepackage{tabularx}
\usepackage{epsfig}
\usepackage{subcaption}
\usepackage{graphicx}
\usepackage{textcomp}
\DeclareCaptionLabelFormat{cont}{#1~#2\alph{ContinuedFloat}}
\usepackage{threeparttable}
\usepackage{array}
\usepackage{lipsum}         % For generating dummy text
\usepackage{wrapfig}
\usepackage{epstopdf}
\usepackage{bm}
\usepackage{makecell}
\usepackage{stfloats}
\usepackage{diagbox}
\usepackage{float}
\usepackage{color}
\usepackage{algorithm}
\usepackage{algorithmicx}
\usepackage{algpseudocode}
\usepackage{tcolorbox}
\usepackage{microtype}      % microtypography
\usepackage{hyperref}
\usepackage{braket}
\usepackage{amscd}
\title{Bridging Visualization and Optimization:\\ Multimodal Large Language Models on Graph-Structured Combinatorial Optimization
}

\author{
  Jie Zhao, Kang Hao Cheong \\
  School of Physical and Mathematical Sciences \\
  Nanyang Technological University \\
  Singapore\\
  \texttt{\{jie.zhao, kanghao.cheong\}@ntu.edu.sg} \\
   \And
  Witold Pedrycz \\
  Department of Electrical and Computer Engineering \\
  University of Alberta \\
  Edmonton, Canada\\
  \texttt{wpedrycz@ualberta.ca} \\
}

\begin{document}
\maketitle

\begin{abstract}
Graph-structured combinatorial challenges are inherently difficult due to their nonlinear and intricate nature, often rendering traditional computational methods ineffective or expensive. However, these challenges can be more naturally tackled by humans through visual representations that harness our innate ability for spatial reasoning. In this study, we propose transforming graphs into images to preserve their higher-order structural features accurately, revolutionizing the representation used in solving graph-structured combinatorial tasks. This approach allows machines to emulate human-like processing in addressing complex combinatorial challenges. By combining the innovative paradigm powered by multimodal large language models (MLLMs) with simple search techniques, we aim to develop a novel and effective framework for tackling such problems. Our investigation into MLLMs spanned a variety of graph-based tasks, from combinatorial problems like influence maximization to sequential decision-making in network dismantling, as well as addressing six fundamental graph-related issues. Our findings demonstrate that MLLMs exhibit exceptional spatial intelligence and a distinctive capability for handling these problems, significantly advancing the potential for machines to comprehend and analyze graph-structured data with a depth and intuition akin to human cognition. These results also imply that integrating MLLMs with simple optimization strategies could form a novel and efficient approach for navigating graph-structured combinatorial challenges without complex derivations, computationally demanding training and fine-tuning.
\end{abstract}

% keywords can be removed

\section{Introduction}
{G}{raph}-structured problems are crucial across various fields due to their ability to model complex relationships \citep{zhang2023large,artime2024robustness,zhang2023theoretical}. In social networks, identifying key nodes can improve information dissemination and marketing strategies \citep{kempe2003maximizing}. Public health also benefits, as targeting influential nodes helps develop effective immunization strategies to prevent disease spread \citep{chen2008finding}. Meanwhile, graph-structured problems are challenging because, unlike traditional Euclidean problems that leverage geometric properties for optimization, graphs are discrete structures lacking clear spatial relationships. This irregularity complicates the application of standard continuous optimization methods. In real-world applications, many graph-structured problems are NP-hard \cite{wang2024asp}. As the number of nodes and edges grows, the combinatorial explosion of possible configurations renders brute-force methods impractical within a reasonable timeframe.

Meta-heuristic algorithms  \citep{gong2016influence,zhao2023self} are effective for complicated problems but face scalability challenges with large datasets. As the problem size increases, the search space expands exponentially, making it harder to find optimal solutions efficiently. Moreover, evaluating solutions is computationally expensive, especially when many iterations are required, further limiting their scalability. Recent years have witnessed incredible progress in the use of graph neural networks (GNNs) on many graph-related tasks \cite{chen2024survey,jin2024survey}, such as node classification \citep{kipf2016semi,velivckovic2017graph} and graph classification \citep{jin2020certified,han2022g}. However, GNNs may lose global structural information due to over-smoothing \citep{chen2020measuring}, where repeated message passing can cause node representations to become indistinguishable, limiting their performance on large-scale networks. In addition, many real-world networks inherently lack labeled data, making it challenging for GNNs to learn meaningful embeddings effectively. Since GNNs are typically trained on specific graph structures, their ability to generalize to unseen networks is limited, further hindering their applicability when applied to various networks. As indicated in a recent study \citep{angelini2023modern}, the performance of modern GNN-based methods is sometimes even worse than simple greedy algorithms, implying that GNNs may not be the optimal backbone for graph-structured combinatorial problems.

Recently, the emergence of large language models (LLMs) has achieved tremendous improvements in many areas such as sentiment analysis \citep{deng2023llms}, translation \citep{gong2024llms}, optimization \citep{romera2024mathematical}, medical applications \citep{chervenak2023promise} and social science \citep{zhang2024toward}, etc. Therefore, it is natural to consider whether the success of LLMs in other fields can be replicated in graph-related tasks \citep{chen2024exploring,tang2024higpt}. As illustrated by \citep{fatemi2023talk,wang2024can}, LLMs are not good at understanding graph-structured data and cannot even deliver acceptable results on some basic tasks. Moreover, LLMs' performance drops drastically with the increase in the graph size. Consequently, it is unlikely that LLMs can directly tackle complex problems in real-world networks at the present stage. 

\begin{figure*}[htbp]
\centering
\includegraphics[height=3.3cm,width=13.9cm]{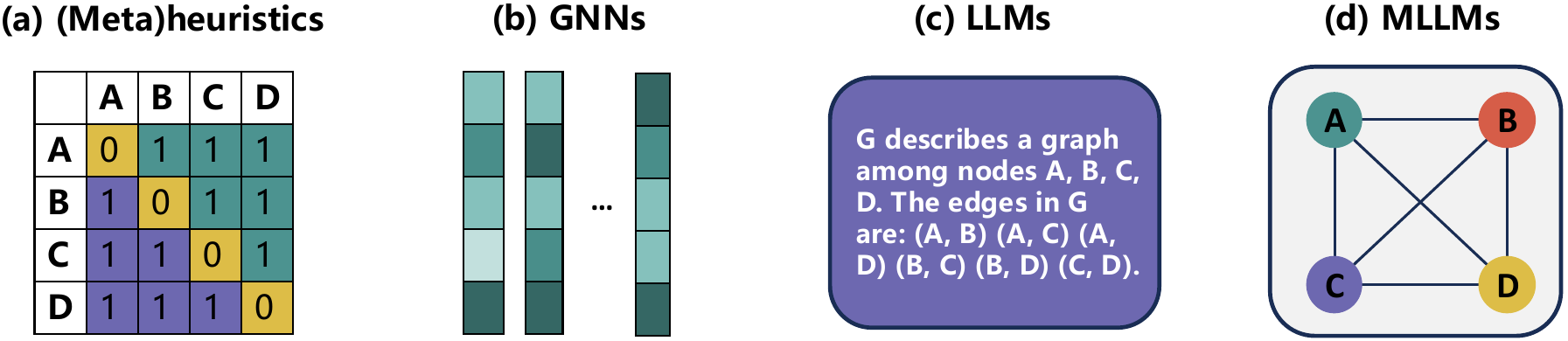}
\caption{The representation of different eras of graph structure. (a) Adjacency matrix; (b) Embedding; (c) Text; (d) Image.} 
\label{era}
\end{figure*}

Over time, the representation of graph-structured data has evolved significantly with the development of computational techniques, as illustrated in Figure \ref{era}. Initially, (meta)heuristic methods focus on directly manipulating graph data through adjacency matrices. Representation learning progressed significantly, as demonstrated by GNNs, which utilize low-dimensional vector spaces to capture the structural properties of graphs, enabling more complex computations.
In the era of LLMs, the fundamental way of representing graph-structured data shifted to natural language, allowing machines to interpret and analyze graphs through textual descriptions. However, graphs are inherently spatial constructs, where the placement, distance, and connections reveal abundant information about the system's structure. Converting a graph into non-visual formats such as adjacency matrices, texts, or embeddings will obscure and lose some structural details, particularly global and high-order information.

In fact, certain problems that are highly complex for machines may be far less challenging for humans, a phenomenon particularly evident in combinatorial optimization. When graph data is properly visualized, humans can use our innate spatial and visual reasoning to effectively tackle these problems. As the advent of multimodal large language models (MLLMs), we may stand on the brink of a transformative shift in tackling such complex problems. Images, as low-loss (potentially loss-free with advancements in visualization) representations of graph structures, can now be processed by machines, enabling them to directly comprehend and analyze graph data like humans.

In this study, we introduce a novel approach that transforms graphs into images to preserve their essential structural features with precision, revolutionizing how graph-structured combinatorial tasks are addressed. By enabling machines to emulate human-like processing, this method provides a powerful and innovative framework for tackling complex combinatorial challenges. We strategically utilize MLLMs to address a range of challenges, from sequential network dismantling (ND) to influence maximization (IM), to demonstrate their unique strengths in handling graph-structured combinatorial problems. The results are highly promising with MLLMs exhibiting remarkable spatial intelligence and delivering outstanding performance on these complex tasks, all without the need for fine-tuning, suggesting a new era for dealing with graph-structured problems may be approaching. Given their simplicity and effectiveness, MLLMs combined with basic optimization techniques hold great potential as a practical solution for tackling complex graph-structured problems in the future. Furthermore, we explore MLLMs' performance on fundamental graph problems, identifying key factors to their effectiveness. We also discuss potential directions for further unlocking the vast potential of MLLMs in this domain.

In visualization, we tailor the strategies to accommodate different network sizes.  For small networks, we display labels for all nodes in the images provided to the MLLMs, referred to as full-label. For large-scale networks, displaying labels for every node is impractical due to the limited canvas size. In these cases, we selectively label only the nodes most likely to be critical, referred to as partial-label. For the network dismantling problem, we use a simple prompt for the MLLMs and find that it is sufficient to achieve excellent performance, showing the model's inherent spatial intelligence without requiring complex instructions. For influence maximization, we adopt an agent-modeling framework that directs the MLLMs to select seed nodes with varying biases.

The the paper is organized as follows: Section \ref{sec.related work} reviews the work related to our paper. Later, we present the proposed visualization method and local search technique for combinatorial problems in Section \ref{sec.met}. Sections \ref{sec.dismantle} and \ref{sec.IM} cover the method and experimental results of graph dismantling and influence maximization tasks, with the latter being examined separately for the full-label and partial-label cases. Section \ref{sec.basic} investigates the performance of MLLMs on the basic graph-structured tasks and we discuss some prospects and conclude this work in Sections \ref{sec.discussion} and \ref{sec.conslution}, respectively.

\section{Related work}\label{sec.related work}

In this section, we review existing studies on graph-structured combinatorial problems, focusing on influence maximization and network dismantling, as well as recent advancements involving LLMs and MLLMs applied to graph problems.

\textbf{Influence Maximization (IM)} is a computational problem in network science where the goal is to identify a set of key nodes in a network that maximizes the spread of information through the network. By setting a predefined diffusion model, the greedy algorithm \citep{kempe2003maximizing} was employed to iteratively identify the node with the largest influence spread. The Cost-Effective Lazy Forward (CELF) algorithm \citep{leskovec2007cost} was then proposed to significantly reduce computational complexity by leveraging the submodularity of the influence function to avoid unnecessary recalculations. Heuristic methods in influence maximization offer a balance between simplicity and effectiveness, such as degree centrality \citep{freeman2002centrality}, betweenness centrality \citep{freeman1977set}, closeness centrality \citep{wasserman1994social}, eigenvalues \citep{berman1994nonnegative}and PageRank \citep{page1999pagerank}, etc \citep{batagelj2003m}. On the other hand, several meta-heuristics have been proposed based on different bio-inspired evolutionary techniques to solve this complex combinatorial problem due to their flexible representation of solutions and effectiveness \cite{tutsoy2023graph}. Gong \textit{et al.} \citep{gong2016influence} proposed a particle swarm optimization to search for the optimal seed. Other techniques are also explored in this task, such as ant colony \citep{salavati2019identifying}, memetic algorithm \citep{gong2016efficient} and differential evolution \citep{li2022efficient}.

The capability of GNNs has shifted research focus from traditional tasks such as node classification to more complex combinatorial optimization challenges \cite{qi2023blockchain}. For instance, Yu \textit{et al.} redefined the influence maximization problem as a regression task by transforming the adjacency matrix into embeddings via GNNs \citep{yu2020identifying}. More recently, Ling \textit{et al.} introduced DeepIM \citep{ling2023deep}, which seeks to capture the latent representations of seed nodes through end-to-end training.

\textbf{Network Dismantling} refers to identifying the minimal group of nodes whose removal most rapidly leads to the network's fragmentation, as outlined in the optimal percolation problem \citep{artime2024robustness}. A straightforward approach involves targeting nodes based on their centrality measures, with the node degree being a primary metric. This method targets highly connected nodes or hubs \citep{albert2000error,cohen2001breakdown}. Various other heuristic measures of centrality are also applicable for pinpointing these critical nodes. Drawing inspiration from decycling-based techniques, CoreHD focuses on decycling a network by sequentially removing the highest-degree nodes within the 2-core \citep{zdeborova2016fast}. Another approach, known as explosive immunization has been introduced by considering explosive percolation (EP) with strategies to keep network clusters highly fragmented. Additionally, there have been advancements in applying machine learning to network attacks, such as graph dismantling with machine learning (GDM) \citep{grassia2021machine} and FINDER \citep{fan2020finding}.

\textbf{LLMs and MLLMs on graph-structured problems:} LLMs have proven effective in many areas, leading to the question of their applicability to graph-structured data. Chen \textit{et al.} employed LLMs as an enhancer and a predictor, respectively \citep{chen2024exploring}. The LLM-based enhancer augments node features, while the LLM-based predictor directly outputs the classification. A model combining LLMs and graph learning methods named GraphLLM was proposed \citep{chai2023graphllm} to enhance the accuracy of reasoning tasks on the text-attributed graphs (TAGs).

However, TAGs are not prevalent as it is challenging to build the label and textual feature for a huge number of nodes. Thus, these LLM-based work is still not enough to tackle real-world problems where there is only structural information available. Thus, some studies sought to directly encode graph structures into text through different prompt engineering techniques \citep{fatemi2023talk,wang2024can}, enabling LLMs to comprehend and analyze these structures. However, experimental results show that LLMs have significantly limited reasoning capabilities, even with small-scale networks, let alone large-scale real-world networks.

Wei \textit{et al.} \cite{wei2024gita} introduced a framework that systematically converts graphs to images and feeds them into MLLMs for seven fundamental graph reasoning tasks. It provides detailed comparisons with LLMs and GNNs to showcase the advantages of using MLLMs with image representations of graphs with its visual intelligence. Similarly, VisionGraph \cite{li24ab} explored leveraging large multimodal models for graph theory problems in a visual context, establishing a toolchain for eight complex graph problem tasks. Beyond the basic graph problems, there are a few work using MLLMs on combinatorial problems. Huang \textit{et al.} used visual and text information to solve the traveling salesman problem (TSP) \citep{huang2024multimodal}. In the following, Elhenawy \textit{et al.} proposed finding the optimal route with graphical data solely and tested the effectiveness of different MLLMs \citep{elhenawy2024eyeballing}. However, the current work has only verified the limited feasibility of MLLMs in combinatorial optimization, and there remains a significant gap before practical application can be achieved. Firstly, the datasets employed in these studies are relatively small, containing at most fewer than 200 nodes. Secondly, the optimization outcomes do not compare favorably with commonly used benchmarks, indicating that the potential of MLLMs has not been fully realized.

\section{Methodology}\label{sec.met}
In this section, we aim to answer critical questions on the application of MLLMs to combinatorial problems: (1) \textit{How to properly visualize the networks, especially the large-scale networks to make it can be processed by MLLMs?} (2) \textit{How to refine the solution suggested by MLLMs efficiently and effectively?}

\subsection{Visualization}\label{sec.visualization}

\begin{figure*}
\centering
\includegraphics[width=0.9\textwidth]{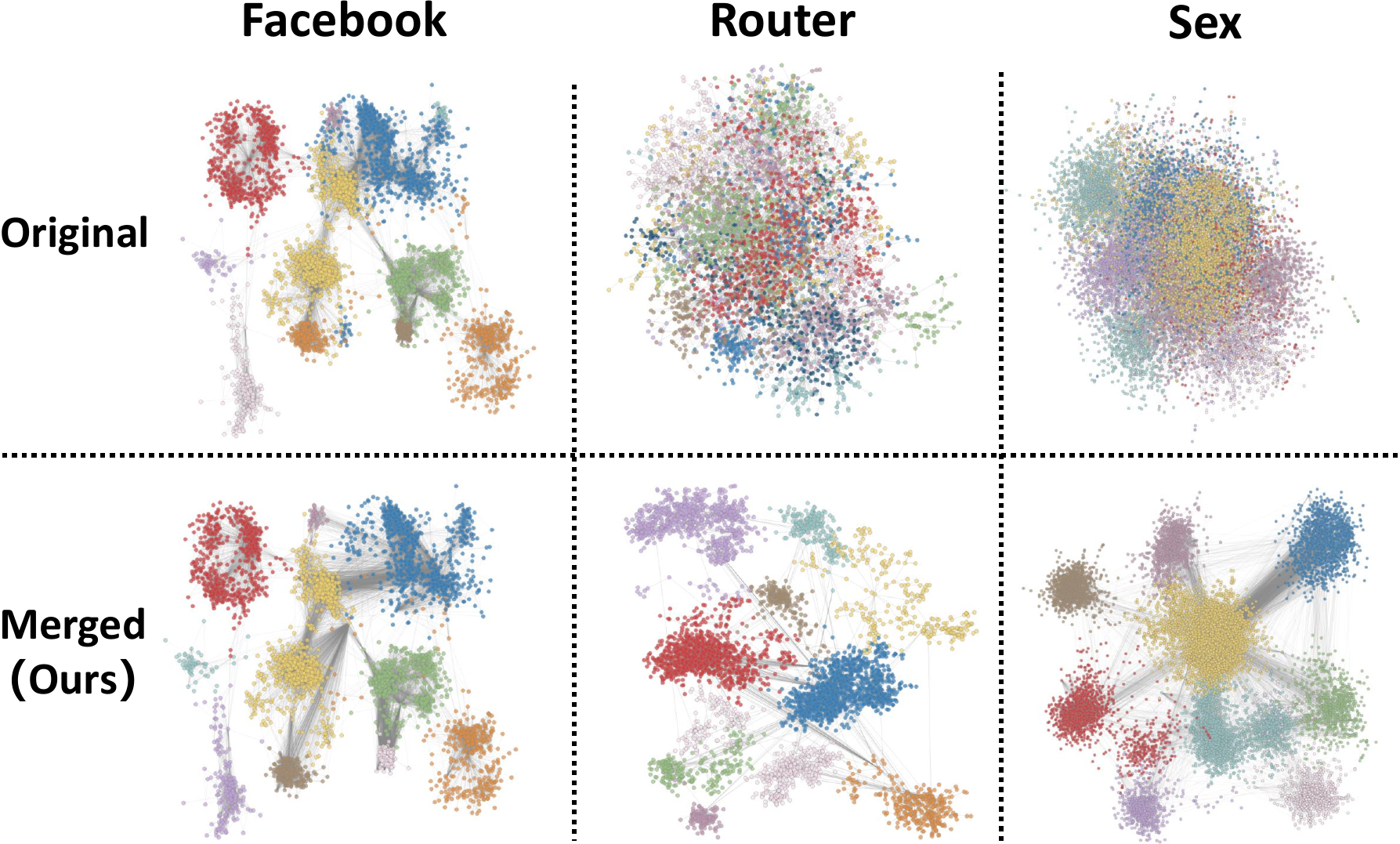}
\caption{Visualization of different networks with their original community structure and corresponding merged structure, each displayed using the Fruchterman-Reingold Layout. The number of original and merged communities for each network is as follows: Facebook (reduced from 13 to 10 communities), Router (reduced from 63 to 10 communities), and Sex (reduced from 170 to 9 communities).}
\label{community_merged}
\end{figure*}

Directly visualizing the network, particularly large-scale networks, on a limited canvas can result in a loss of critical structural information, such as community structures. However, when applying standard community detection algorithms like Fastgreedy \citep{clauset2004finding}, the number of communities detected can often exceed practical utility, especially in large networks. These algorithms tend to identify many small communities that may be of less relevance or too granular for specific applications. In such a case, it is also very difficult to reflect the essential structural information, as shown in the upper side of Figure \ref{community_merged}. Therefore, there is a need for a method to merge these smaller communities into larger, more meaningful groups.

Here, we propose an algorithm to merge small communities into fewer, larger communities while maintaining the integrity and connectivity of the original network structure. The goal is to reduce the number of communities to a more manageable size, aligning with the specific analytical need. That is: Given a graph $G$, an initial set of communities $C$, and a target number of communities $T$, we will merge smaller communities into their nearest neighbors until the number of communities is reduced to $T$. This algorithm is detailed as follows:

\begin{enumerate}
    \item \textbf{Identify the Smallest Community}: In each iteration, the algorithm identifies the smallest community by comparing the sizes of all communities.
    \item \textbf{Count Edges to Other Communities}: For each edge in the graph, the algorithm checks if the edge connects the smallest community to any other community. It keeps track of how many edges each neighboring community has connected to the smallest community.
    \item \textbf{Find the Closest Community}: The community with the highest number of edges connected to the smallest community is chosen as the "closest" community.
    \item \textbf{Merge Communities}: All nodes in the smallest community are reassigned to the closest community. The indices of the other communities are adjusted accordingly to reflect the reduction in the number of communities.
    \item \textbf{Repeat}: This process continues until the number of communities equals the target number.
\end{enumerate}

To ensure that important nodes do not overlap and can be easily recognized by MLLMs, it is necessary to position top-ranked nodes (e.g., those with high degrees) farther apart, while other nodes are arranged closer to the community centroid. To this end, we propose a method to adjust the positions of nodes in a graph \( G \) according to their community structure and layout style, with a specific emphasis on spatial differentiation of influential nodes. Each node's initial coordinates are computed by a graph layout algorithm. Given a community structure \( \mathcal{C} = \{\mathcal{C}_1, \mathcal{C}_2, \ldots, \mathcal{C}_k\} \), where each \( \mathcal{C}_i \) represents a set of nodes belonging to the same community, the centroid \( \mathbf{c}_i \) of community \( \mathcal{C}_i \) is calculated as:
\[
\mathbf{c}_i = \frac{1}{|\mathcal{C}_i|} \sum_{v \in \mathcal{C}_i} \mathbf{p}_v,
\]
where \( \mathbf{p}_v \) denotes the position of node \( v \). Nodes within each community \( \mathcal{C}_i \) are ranked based on their degrees, identifying the top-N highest-degree nodes (denoted as \( T_i \subset \mathcal{C}_i \) with cardinality \(|T_i| = N\)) for further adjustment.

For each node \( v \in \mathcal{C}_i \), a new position \( \mathbf{p}'_v \) is calculated to reflect its distance from the community centroid \( \mathbf{c}_i \), controlled by an adjustment parameter \( d \). Specifically, 

$\bullet$ For nodes in \( T_i \), the adjustment distance remains unchanged. 

$\bullet$ For nodes in \( \mathcal{C}_i \setminus T_i \), the adjustment distance is scaled proportionally by a factor of \( 1 - d \), that is
\[
\mathbf{p}'_v = \mathbf{p}_v \cdot d + \mathbf{c}_i \cdot (1 - d),
\]
where parameter $d \in (0,1)$ controls the spatial separation within a community, where a lower $d$ creates a denser layout and a higher $d$ spreads nodes farther apart.

% \begin{table}[ht]
% \centering
% \caption{The number of original communities and the merged communities.} 
% \label{tab.Num Com}
% \begin{tabular}{lcccccc}
% \Xhline{5\arrayrulewidth} 
% \textbf{Network} & \textbf{Original} & \textbf{Merged}\\
% \hline
% Facebook & 13 & 9 \\
% Router & 63 & 9  \\
% Sex & 170 & 10 \\
% \Xhline{5\arrayrulewidth}
% \end{tabular}
% \end{table}

\subsection{Local search}\label{sec.local search}
To further enhance the influence spread of a seed set initially suggested by MLLMs, we propose a local search method: It iteratively attempts to improve the seed set by exploring replacements: for each node in the current seed set, and the algorithm examines its neighbors (sorted by degree or betweenness) to find a suitable candidate for replacement. If replacing a node with a top-ranked neighbor increases the influence spread, the seed set is updated. The process continues until no further improvements can be made or the maximum iteration is reached. The pseudocode of the process can be found in Algorithm \ref{local_search}. The influence spread is evaluated using a predefined influence diffusion model, such as the Independent Cascade (IC) and Linear Threshold (LT) models. For the sake of efficiency, the iteration number is set to 5 and the simulation number of the spreading process is 5,000.

\begin{algorithm}
\caption{Local Search Algorithm for Influence Maximization}
\begin{algorithmic}[1]
\State \textbf{Precompute} degrees and betweenness for all nodes in the graph.
\State \textbf{Initialize} seed set \( S \) with MLLM to \( \text{InitialSeed} \).
\State \textbf{Evaluate} the initial influence spread of \( S \) based on the model, stored in \( \text{BsetSpread} \).
\For{\( \text{MaxIter} \) iterations}
    \State \( \text{improved} \leftarrow \text{False} \)
    \For{each node \( v \) in \( S \)}
        \State \( N(v) \leftarrow \) list of neighbors of \( v \).
        \State Sort \( N(v) \) based on degree or betweenness, randomly chosen.
        \State \( u \leftarrow \) top-ranked neighbor not in \( S \).
        \State \( S' \leftarrow (S \setminus \{v\}) \cup \{u\} \).
        \State Calculate \( \text{NewSpread} \) for \( S' \) using the selected model.
        \If{\( \text{NewSpread} > \text{BsetSpread} \)}
            \State \( S \leftarrow S' \).
            \State \( \text{BsetSpread} \leftarrow \text{NewSpread} \).
            \State \( \text{improved} \leftarrow \text{True} \).
            \State \textbf{break}
        \EndIf
    \EndFor
    \If{not \( \text{improved} \)}
        \State \textbf{break}
    \EndIf
\EndFor
\end{algorithmic}
\label{local_search}
\end{algorithm}

\section{Influence maximization}\label{sec.IM}

\textbf{Influence Maximization (IM)} aims to find a subset of seed nodes \( S \subset V \) that maximizes the overall influence spread across a network. This spread is governed by a probabilistic diffusion model. The goal of the problem is to maximize $ \sigma(S)$ where \( \sigma(S) \) denotes the expected spread of influence starting from the seed set \( S \).

% % Independent Cascade (IC) Model
% \textbf{Independent Cascade (IC) model}: It is a diffusion model used to simulate the spread of influence in a network. In this model, each activated node has a single chance to activate each of its inactive neighbors with a given probability \( p \). If node \( u \) becomes active at time \( t \), it will attempt to activate each inactive neighbor \( v \) at time \( t+1 \). The process continues until no more activations are possible. 

% % Linear Threshold (LT) Model
% \textbf{Linear Threshold (LT) model}: It is another diffusion model that assumes each node in the network has a threshold \(\theta_v \in [0, 1]\). A node \( v \) becomes active if the sum of the influences from its active neighbors exceeds its threshold. Each edge \((u, v)\) has an associated weight \( w_{uv} \) such that \(\sum_{u \in N(v)} w_{uv} \leq 1\), where \( N(v) \) is the set of neighbors of \( v \). The activation condition for node \( v \) is:
% \[ \sum_{u \in N(v), \text{active}} w_{uv} \geq \theta_v. \]

\subsection{Benchmarks}\label{sec.benchmark}
\textbf{Degree} measures the number of direct connections a node has. For a node \( v \), degree centrality \( \text{DC}(v) \) is given by:
$$
\text{DC}(v) = \sum_{u \in V} a_{vu},
$$
where \( a_{vu} \) is the element of the adjacency matrix indicating the presence of an edge between nodes \( v \) and \( u \).

\textbf{Betweenness} measures the extent to which a node lies on the shortest paths between other nodes. For a node \( v \), betweenness centrality \( \text{BC}(v) \) is given by:
$$
\text{BC}(v) = \sum_{s \neq v \neq t} \frac{\sigma_{st}(v)}{\sigma_{st}},
$$
where \( \sigma_{st} \) is the total number of shortest paths from node \( s \) to node \( t \), and \( \sigma_{st}(v) \) is the number of those paths that pass through \( v \).

\textbf{Closeness} measures how close a node is to all other nodes in the network. For a node \( v \), closeness centrality \( \text{CC}(v) \) is given by:
$$
\text{CC}(v) = \frac{1}{\sum_{u \in V} d(v, u)},
$$
where \( d(v, u) \) is the shortest distance between nodes \( v \) and \( u \).

\textbf{PageRank} measures the influence of a node based on the idea that connections to high-scoring nodes contribute more to the score of the node. For a node \( v \), it is evaluated by:
$$
\text{PR}(v) = \frac{1 - \alpha}{|V|} + \alpha \sum_{u \in \text{N}_{(v)}} \frac{\text{PR}(u)}{out(u)},
$$
where $\alpha$ is a damping factor and $\text{N}_{(v)}$ is node $v$'s neighbors.

\textbf{Collective influence (CI)} of a node at distance $l$ is determined by taking into account both the degree of the node and the degrees of nodes that are $l$ steps away \citep{morone2015influence}. Specifically, the CI of a node $v$ in a network is defined as:

$$
CI_l(v) = (k_v - 1) \sum_{u \in \partial B_l(v)} (k_u - 1),
$$
where \( k_v \) is the degree of the node \( v \) and
\( \partial B_l(v) \) represents the set of nodes that are exactly \( l \) steps away from \( v \) (the boundary of the ball of radius \( l \) around \( v \)).
\( k_u \) is the degree of a node \( u \) in the boundary set.

\textbf{DeepIM} is a GNN-based framework that models the seed set's representation within a latent space. This representation is concurrently trained with a model that comprehends the fundamental network diffusion mechanism with an end-to-end training approach \citep{ling2023deep}.

\subsection{Experimental setting}\label{sec.experimental setting}
As our work is not aiming to compare the performance of MLLMs but to explore a novel solution to graph tasks, we directly select the state-of-the-art model \textit{gpt-4o-2024-08-06} as our backbone. In network dismantling, the agent makes 20 attempts on each network. The structural details of the analyzed networks are presented in Table \ref{tab.topological}. Networks with fewer than 150 nodes are classified as small networks, while those with 150 or more nodes are classified as large networks. For influence maximization, we design 4 agents for the partial-label case and 3 agents for the full-label case, with each agent sampling nodes 10 times. In the validation, we use the Monte Carlo method to simulate 100,000 spreading processes for the IC and LT models. The infection probability of the IC model is set to 0.1. The effectiveness of influence maximization of different methods is examined with two spreading models Independent Cascade model \citep{robson2024cynetdiff} and Linear Threshold model \citep{riquelme2018centrality}. In the following experiment, \textbf{MLLM} refers to the best seeds among all attempts of agents and \textbf{MLLM-ls} refers to the best seeds among all attempts of agents after local search. The prompt of influence maximization is shown in Table \ref{table:prompts}.

\begin{table*}[ht]
\centering
\caption{The structural information of tested real-world networks after removing self-loops and isolated components. $|\mathcal{V}|$ and $|\mathcal{E}|$ refer to the number of nodes and edges, respectively.} 
\label{tab.topological}
\begin{tabular}{cccccccc}
\Xhline{5\arrayrulewidth} 
\textbf{Network} & \textbf{Karate} & \textbf{Dolphins} & \textbf{Lesmis} & \textbf{Polbooks} & \textbf{Facebook} & \textbf{Router} & \textbf{Sex} \\ 
\hline
$|\mathcal{V}|$ & 34    & 62    & 77   & 105 & 4,039   & 5,022&15,810 \\
$|\mathcal{E}|$ & 78    & 159   & 254  & 441 & 88,234  &6,258& 35,840 \\
\Xhline{5\arrayrulewidth}
\end{tabular}
\end{table*}

\begin{table*}[h!]
\centering
\caption{Context-setting and output directive prompts for influence maximization. Context-setting is placed at the beginning of the prompt to explain the input information and the played role to agents and the output is placed at the end of the prompt to restrict the output format.}
\label{table:prompts}
\begin{tabular}{m{3.5cm}m{5.5cm}m{5.5cm}}
\Xhline{5\arrayrulewidth}
\textbf{Task} & \textbf{Context-setting prompt} & \textbf{Output directive prompt} \\ \hline
\textbf{Influence maximization (full label)} & You are an expert in network science and will be provided with one network in the form of an image. & Do NOT output any other text or explanation. Just tell me the node IDs only. Your answer should be only a list as [node\_id, ..., node\_id] \\ \hline
\textbf{Influence maximization (partial label)} & You are an expert in network science and will be provided with one network in the form of an image. The network is divided into different communities and the nodes in the same community are of the same color. & Do NOT output any other text or explanation. Just tell me the node IDs only. Your answer should be only a list as [node\_id, ..., node\_id] \\ \Xhline{5\arrayrulewidth}
\end{tabular}
\end{table*}
\subsection{Small-scale network}

In this section, we employ an agent-based method for IM. Each agent is equipped with unique criteria. The visualization method and agent vary with network sizes. Here, seed nodes in IM are selected simultaneously, introducing additional challenges: (1) MLLMs must account for the global pattern and interconnections among seeds; (2) The selected seeds must satisfy specific requirements, such as seed size, and ensure no repetition.

\begin{figure*}
\centering
\includegraphics[width=0.8\textwidth]{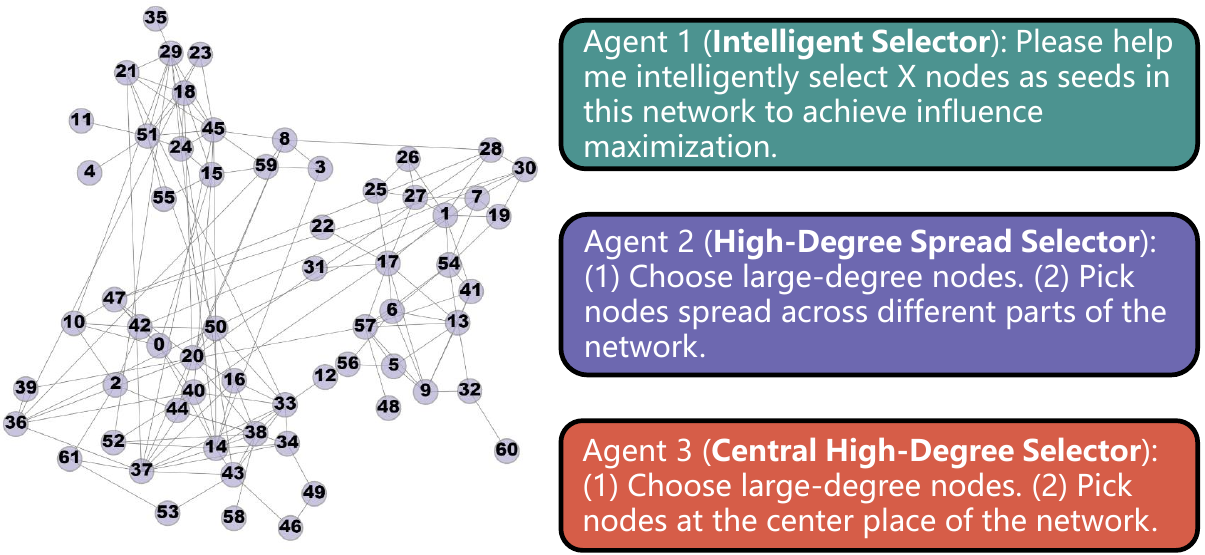}
\caption{The illustrations of three agents for IM on small-scale networks. The full-label network (left) will be inputted into MLLM along with the prompts for agents (right).}
\label{IM(s)-diagram}
\end{figure*}

Figure \ref{IM(s)-diagram} shows the MLLM-based IM in small-scale networks where all nodes are visualized on a single canvas with labeled node IDs. The full-label network will be input to MLLM as an image for multiple-node selection in one go. We design each agent focusing on a different criterion. Agent 1 solely relies on the intelligence of MLLM while Agents 2 and 3 are equipped with specific hints, focusing on the distributed and central parts, respectively. The prompt for Agent 1 is also placed in front of the prompt for the other agents as the leading sentence to explain the task.

\begin{table*}[]
\centering
\captionof{table}{The validations across different networks and MLLM agents. Three validations are included:  (1) the ratio of seed nodes correctly matching the specified seed size,  (2) the ratio of seed nodes that correctly exclude non-existent nodes, and (3) the ratio of non-redundant seed nodes in each seed set.}
\label{validation_small}
\begin{tabular}{ccccccccccc}  
    \Xhline{5\arrayrulewidth}  
    \multirow{2}{*}{\textbf{Dolphins}}&  
    \multicolumn{2}{c}{\textbf{Agent 1}}&\multicolumn{2}{c}{\textbf{Agent 2}}&\multicolumn{2}{c}{\textbf{Agent 3}}\cr  
    \cmidrule(lr){2-3} \cmidrule(lr){4-5} \cmidrule(lr){6-7} 
    & $\boldsymbol{|S| = 5}$& $\boldsymbol{|S|=10}$& $\boldsymbol{|S| = 5}$& $\boldsymbol{|S|=10}$& $\boldsymbol{|S| = 5}$& $\boldsymbol{|S|=10}$& \cr 
    \midrule   
    Validation 1 & 100.0\% & 100.0\% & {100.0\%} & {100.0\%} & 100.0\% & {100.0\%}  \\
    Validation 2 & 100.0\% & 100.0\% & {100.0\%} & {100.0\%} & 100.0\% & {100.0\%}  \\
    Validation 3 & 100.0\% & 100.0\% & {100.0\%} & {100.0\%} & 100.0\% & {100.0\%}  \\
    \Xhline{5\arrayrulewidth}  
    \multirow{2}{*}{\textbf{Lesmis}}&  
    \multicolumn{2}{c}{\textbf{Agent 1}}&\multicolumn{2}{c}{\textbf{Agent 2}}&\multicolumn{2}{c}{\textbf{Agent 3}}\cr  
    \cmidrule(lr){2-3} \cmidrule(lr){4-5} \cmidrule(lr){6-7} 
    & $\boldsymbol{|S| = 5}$& $\boldsymbol{|S|=10}$& $\boldsymbol{|S| = 5}$& $\boldsymbol{|S|=10}$& $\boldsymbol{|S| = 5}$& $\boldsymbol{|S|=10}$& \cr 
    \midrule   
    Validation 1 & 100.0\% & 100.0\% & {100.0\%} & {100.0\%} & 100.0\% & {100.0\%}  \\
    Validation 2 & 100.0\% & 100.0\% & {100.0\%} & {100.0\%} & 100.0\% & {100.0\%}  \\
    Validation 3 & 100.0\% & 100.0\% & {100.0\%} & {100.0\%} & 100.0\% & {99.0\%}  \\
    \Xhline{5\arrayrulewidth}
    \multirow{2}{*}{\textbf{Polbooks}}&  
    \multicolumn{2}{c}{\textbf{Agent 1}}&\multicolumn{2}{c}{\textbf{Agent 2}}&\multicolumn{2}{c}{\textbf{Agent 3}}\cr  
    \cmidrule(lr){2-3} \cmidrule(lr){4-5} \cmidrule(lr){6-7} 
    & $\boldsymbol{|S| = 5}$& $\boldsymbol{|S|=10}$& $\boldsymbol{|S| = 5}$& $\boldsymbol{|S|=10}$& $\boldsymbol{|S| = 5}$& $\boldsymbol{|S|=10}$& \cr 
    \midrule  
    Validation 1 & 100.0\% & 100.0\% & {100.0\%} & {100.0\%} & 100.0\% & {100.0\%}   \\
    Validation 2 & 100.0\% & 100.0\% & {100.0\%} & {100.0\%} & 100.0\% & {100.0\%}   \\
    Validation 3 & 100.0\% & 100.0\% & {100.0\%} & {100.0\%} & 100.0\% & {100.0\%}   \\
    \Xhline{5\arrayrulewidth}   
\end{tabular}
\end{table*}

\textbf{MLLM agents are capable of selecting seed sets that align with the specified criteria in the full-label case:} Due to the LLM hallucination \citep{xu2024hallucination,duan2024llms}, we examine the feasibility and correctness of selected seeds by MLLM. The criteria include checking for repetitive or invalid nodes in the seed nodes and ensuring that the selected seed size meets our specifications. Table \ref{validation_small} shows that across three networks, the validation results are consistently high, with most metrics achieving 100\% accuracy for all agents.

\begin{figure}
  \centering
  \includegraphics[width=0.8\textwidth]{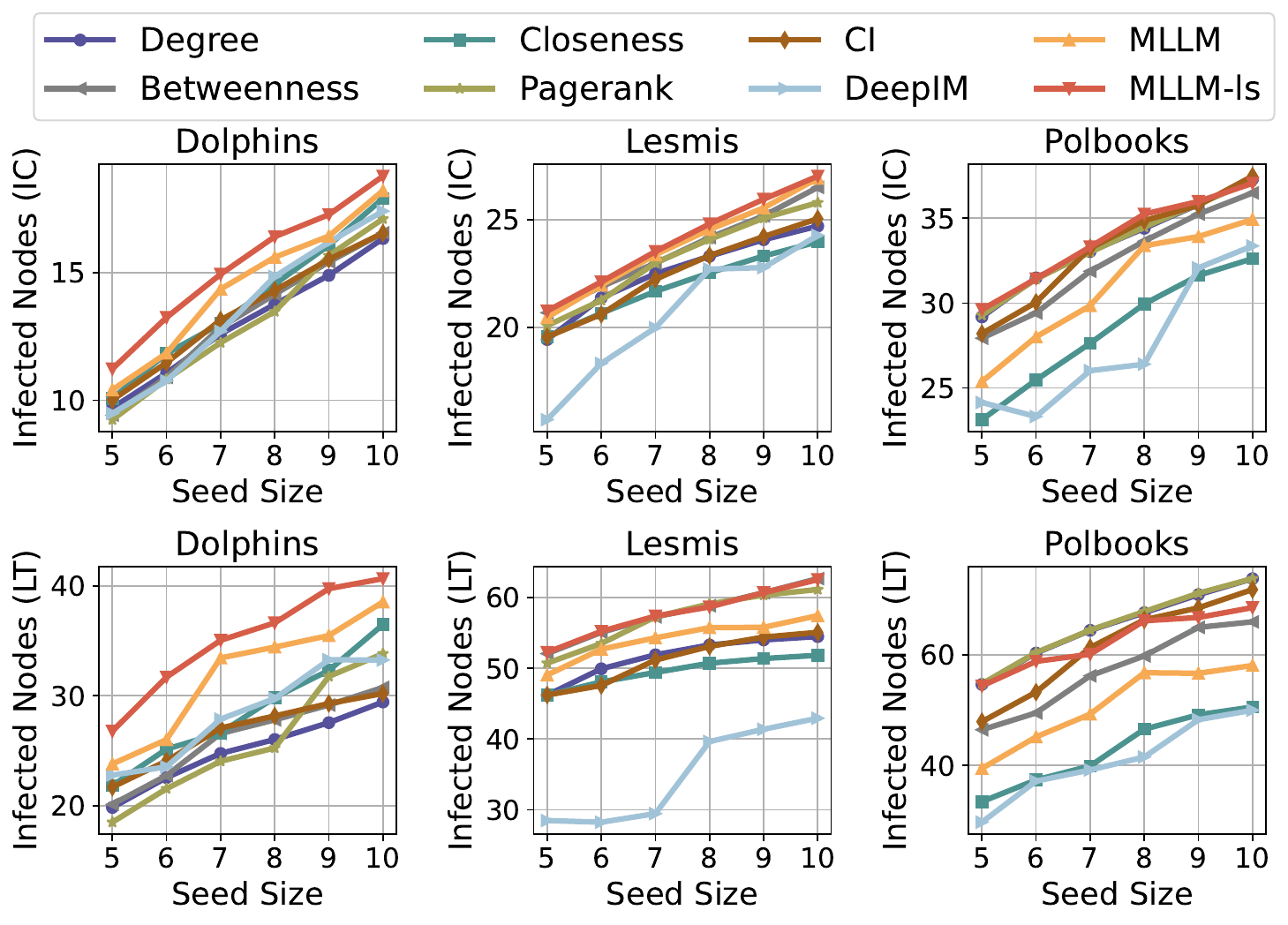}
  \caption{The comparative IM performance on small-scale networks with the IC and LT models.}
  \label{IM_small}
\end{figure}

\textbf{MLLM plus local search would become a new paradigm for combinatorial optimization:} Figure \ref{IM_small} shows the results of IM using various strategies. In both IC and LT models, the MLLM-ls consistently outperforms other strategies, achieving a higher number of infected nodes across all seed sizes compared to traditional centrality methods such as degree, betweenness, and CI, as well as representation learning-based DeepIM, in selecting seeds for IM within networks. As shown in Figure \ref{IM(s)-diagram}, the agents' prompts are straightforward and intuitive, highlighting that MLLM is not only effective but also user-friendly, making it highly accessible for practical use.

\begin{figure}
  \centering
  \includegraphics[width=0.8\textwidth]{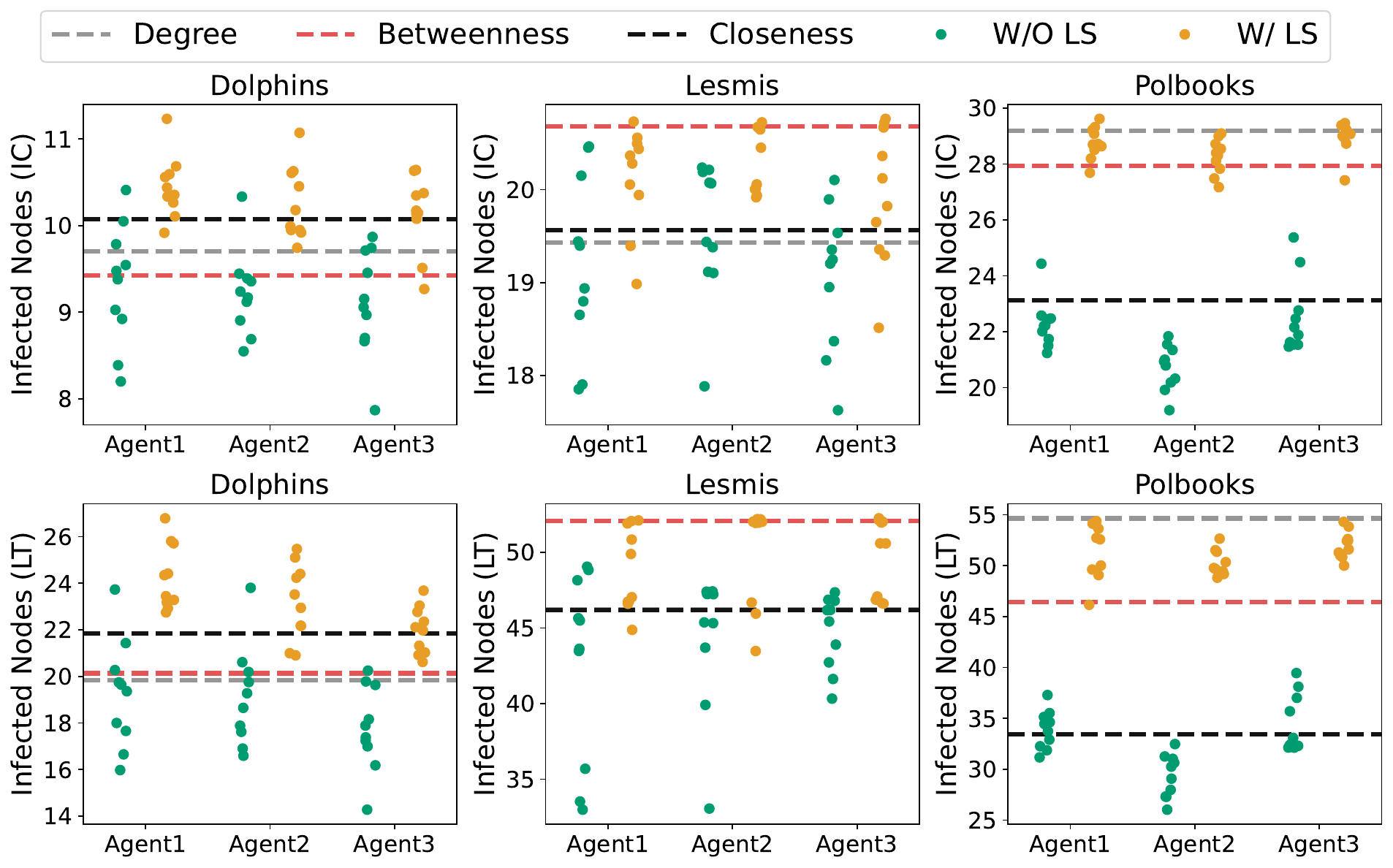}
  \caption{The IM result of MLLM agents with and without local search on small-scale networks. LS refers to local search.}
  \label{Distribution_small}
\end{figure}

\textbf{MLLM exhibits an excellent inherent intelligence:} Figure \ref{Distribution_small} shows the distribution of infected nodes using different seed nodes suggested by different agents. The performance of the different agents across networks varies significantly due to their distinct strategies. Agent 1, which operates without specific hints, consistently performs as well as other agents with guidance across all networks. This indicates that the MLLM's capability has reached a high level of intelligence and can make optimal selections, even without explicit guidance.

\textbf{The visualization poses a challenge to MLLM for accurately recognizing the node in the dense network:} In the Polbooks network, which is both larger and denser, the visual complexity increases, making it more challenging for the agents to effectively recognize optimal seed nodes. This is where local search plays a crucial role, as demonstrated by the improvement on Polbooks as well as Dolphins and Lesmis. It helps refine the selection in a visually dense network, where visual inspection alone may not be sufficient.

\subsection{Large-scale network}
\begin{figure*}
\centering
\includegraphics[width=0.98\textwidth]{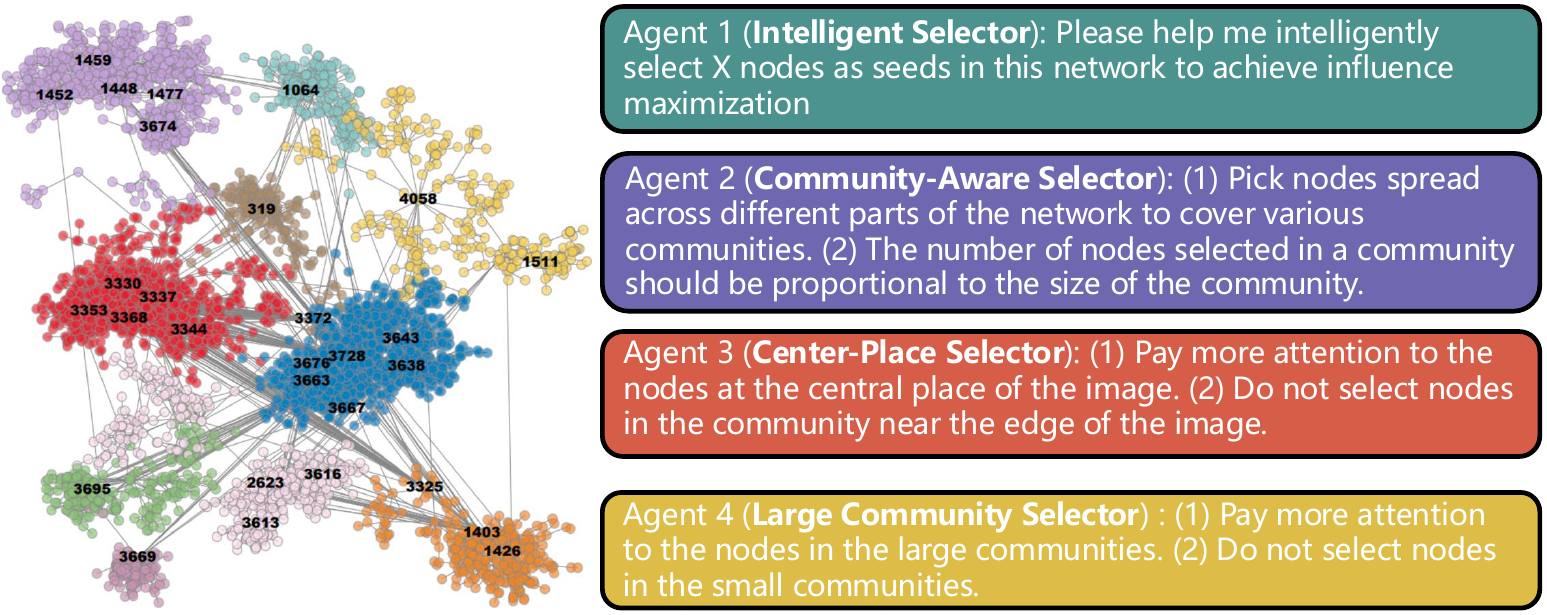}
\caption{The illustrations of MLLM-based IM on large-scale networks. The partial-label network (left) will be inputted into MLLM along with the prompts for agents (right).}
\label{IM-diagram}
\end{figure*}

The details of agents for the large-scale networks are shown in Figure \ref{IM-diagram}. Due to the substantial number of nodes of large-scale networks, it is impractical to plot all the labels in a canvas of limited size. Thus, only a certain ratio of high-degree nodes of each network is displayed in the image.

In this case, the input to MLLM becomes an image with partial labels. As seen from the prompt for agents and the input image, we also include the community information compared to the full-label case. This is because (1) While MLLM shows strong spatial intelligence, we still need some assistance to explicitly guide it in selecting area nodes when incorporating selection biases. (2) There is still a lack of visualization tools that effectively display the network structure globally. Thus, we utilize community detection to cluster densely connected nodes and separate loosely connected parts for better visualization. Advancements in visualization will unlock significant potential for MLLM in large-scale graph-structured problems, which will be discussed further in Section \ref{sec.discussion}.

\begin{table*}[ht]
\centering
\caption{The validations across different networks and MLLM agents. Three validations are included:  (1) the ratio of seed nodes correctly matching the specified seed size,  (2) the ratio of seed nodes that correctly exclude non-existent nodes, and (3) the ratio of non-redundant seed nodes in each seed set.}
\label{Validation_large}
\begin{subtable}{\textwidth}
    \centering
    \begin{tabular}{ccccccccccc}  
    \Xhline{5\arrayrulewidth}  
    \multirow{2}{*}{\textbf{Facebook}}&  
    \multicolumn{2}{c}{\textbf{Agent 1}}&\multicolumn{2}{c}{\textbf{Agent 2}}&\multicolumn{2}{c}{\textbf{Agent 3}}&\multicolumn{2}{c}{\textbf{Agent 4}}\cr  
    \cmidrule(lr){2-3} \cmidrule(lr){4-5} \cmidrule(lr){6-7}\cmidrule(lr){8-9} 
    & $\boldsymbol{|S| = 10}$& $\boldsymbol{|S|=20}$& $\boldsymbol{|S| = 10}$& $\boldsymbol{|S|=20}$& $\boldsymbol{|S| = 10}$& $\boldsymbol{|S|=20}$& $\boldsymbol{|S| = 10}$& $\boldsymbol{|S|=20}$\cr 
    \midrule   
    Validation 1 & 100.0\% & 100.0\% & {100.0\%} & {100.0\%} & 100.0\% & {100.0\%} & {100.0\%} & {100.0\%} \\
    Validation 2 & {99.0\%} & {97.5\%} & {99.0}\% & 98.5\% & 98.0\% & {99.0\%} & {99.0\%} & 100.0\% \\
    Validation 3 & 100.0\% & 99.5\% & 100.0\% & 97.5\% & {100.0\%} & 99.5\% & 100.0\% & 99.0\% \\
    % \Xhline{5\arrayrulewidth}   
    \end{tabular}
\end{subtable}

% \vspace{0.5em} % Adds some vertical space between subtables

\begin{subtable}{\textwidth}
    \centering
    \begin{tabular}{ccccccccccc}  
    \Xhline{5\arrayrulewidth}  
    \multirow{2}{*}{\textbf{Router}}&  
    \multicolumn{2}{c}{\textbf{Agent 1}}&\multicolumn{2}{c}{\textbf{Agent 2}}&\multicolumn{2}{c}{\textbf{Agent 3}}&\multicolumn{2}{c}{\textbf{Agent 4}}\cr  
    \cmidrule(lr){2-3} \cmidrule(lr){4-5} \cmidrule(lr){6-7}\cmidrule(lr){8-9} 
    & $\boldsymbol{|S| = 10}$& $\boldsymbol{|S|=20}$& $\boldsymbol{|S| = 10}$& $\boldsymbol{|S|=20}$& $\boldsymbol{|S| = 10}$& $\boldsymbol{|S|=20}$& $\boldsymbol{|S| = 10}$& $\boldsymbol{|S|=20}$\cr 
    \midrule   
    Validation 1 & 100.0\% & 100.0\% & {100.0\%} & {100.0\%} & 100.0\% & {100.0\%} & {100.0\%} & {100.0\%} \\
    Validation 2 & {98.0\%} & {98.5\%} & {99.0}\% & 98.5\% & 98.0\% & {91.5\%} & {98.0\%} & 96.0\% \\
    Validation 3 & 100.0\% & 99.5\% & 100.0\% & 97.5\% & {100.0\%} & 99.5\% & 100.0\% & 99.0\% \\
    % \Xhline{5\arrayrulewidth}   
    \end{tabular}
\end{subtable}

% \vspace{0.5em} % Adds some vertical space between subtables

\begin{subtable}{\textwidth}
    \centering
    \begin{tabular}{ccccccccccc}  
    \Xhline{5\arrayrulewidth}  
    \multirow{2}{*}{\textbf{Sex}}&  
    \multicolumn{2}{c}{\textbf{Agent 1}}&\multicolumn{2}{c}{\textbf{Agent 2}}&\multicolumn{2}{c}{\textbf{Agent 3}}&\multicolumn{2}{c}{\textbf{Agent 4}}\cr  
    \cmidrule(lr){2-3} \cmidrule(lr){4-5} \cmidrule(lr){6-7}\cmidrule(lr){8-9} 
    & $\boldsymbol{|S| = 10}$& $\boldsymbol{|S|=20}$& $\boldsymbol{|S| = 10}$& $\boldsymbol{|S|=20}$& $\boldsymbol{|S| = 10}$& $\boldsymbol{|S|=20}$& $\boldsymbol{|S| = 10}$& $\boldsymbol{|S|=20}$\cr 
    \midrule   
    Validation 1 & 100.0\% & 100.0\% & {100.0\%} & {100.0\%} & 100.0\% & {100.0\%} & {100.0\%} & {100.0\%} \\
    Validation 2 & {93.0\%} & {85.0\%} & {89.0}\% & 88.0\% & 92.0\% & {91.5\%} & {92.0\%} & 80.0\% \\
    Validation 3 & 99.0\% & 99.5\% & 99.0\% & 99.5\% & {100.0\%} & 99.5\% & 99.0\% & 97.5\% \\
    \Xhline{5\arrayrulewidth}   
    \end{tabular}
\end{subtable}
\end{table*}

As seen in Table \ref{Validation_large}, the agents demonstrate strong correctness across most networks, particularly in correctly matching the specified seed size and avoiding selecting redundant nodes. The displayed nodes in Sex are more than the other two networks due to its larger size, which poses a challenge to accurately identifying the node label, reflected by the relatively low accuracy in Validation 2. A further discussion can be found in Section \ref{sec.discussion} and Figure \ref{recognition}(c).

\begin{figure}
  \centering
\includegraphics[width=0.8\textwidth]{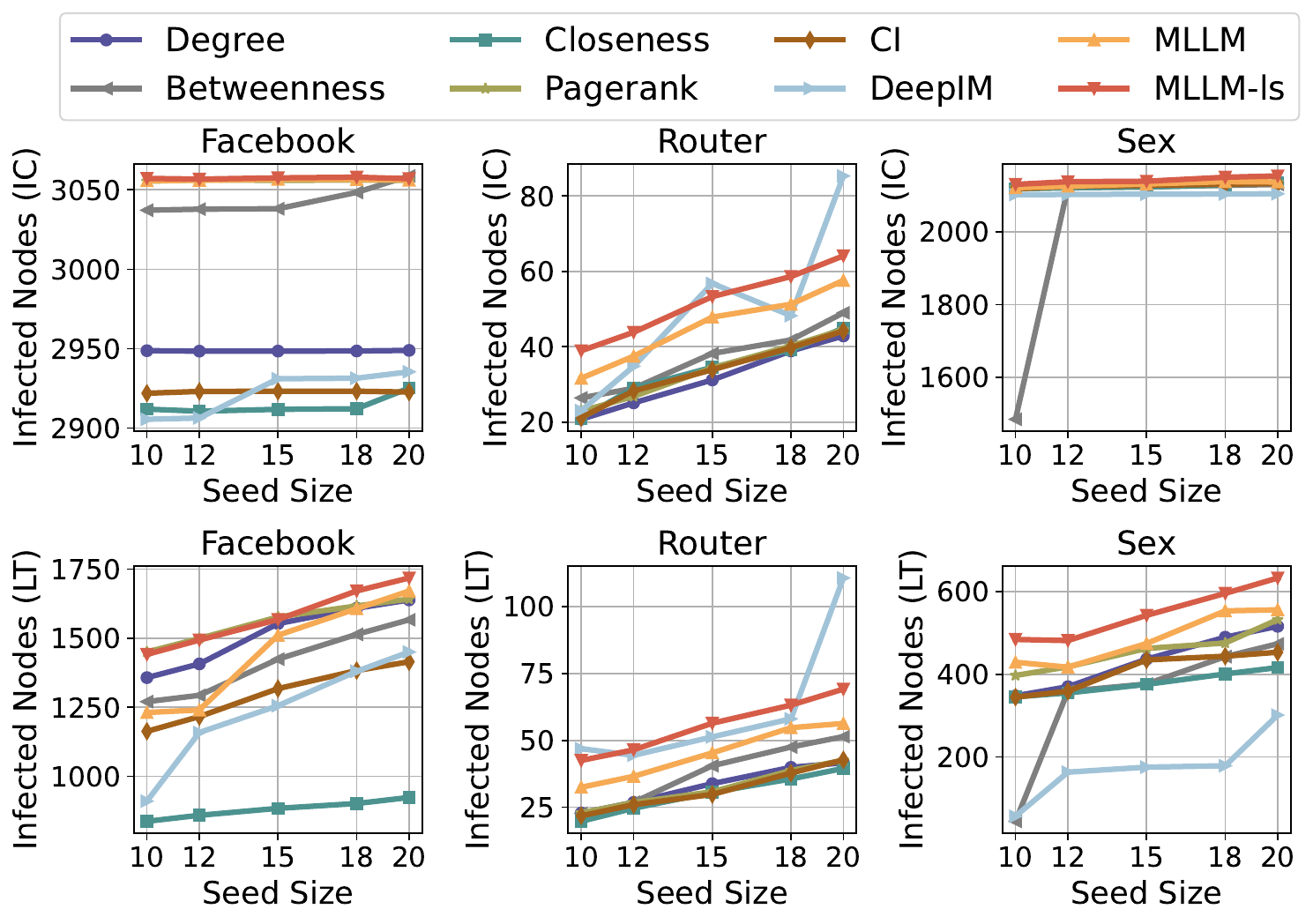}
\caption{The comparative IM performance on large-scale networks with the IC and LT models.}
\label{IM_large}  
\end{figure}

\textbf{MLLM performs also well on large-scale networks}: Figure \ref{IM_large} presents the IM results on large-scale networks. As observed, MLMM-ls outperforms all tested methods including the state-of-the-art GNN-based DeepIM, while MLLM without local search can also surpass most centrality and hand-crafted approaches, suggesting the applicability of MLLM on real-world networks that are typically large-scale. Considering its simplicity and effectiveness, MLLM along with basic optimization techniques will be a promising candidate for large-scale graph problems.

Figure \ref{distribution_large} shows the IM results of different agents on large-scale networks. In several cases, the MLLM agents, particularly Agent 1, outperform traditional centrality-based methods such as degree or betweenness centrality, even when local search is not applied. This suggests that MLLMs have an inherent capability to select influential nodes even without being explicitly directed, rivaling or exceeding conventional metrics that rely on predefined structural properties.

\begin{figure}
\centering
\includegraphics[width=0.8\textwidth]{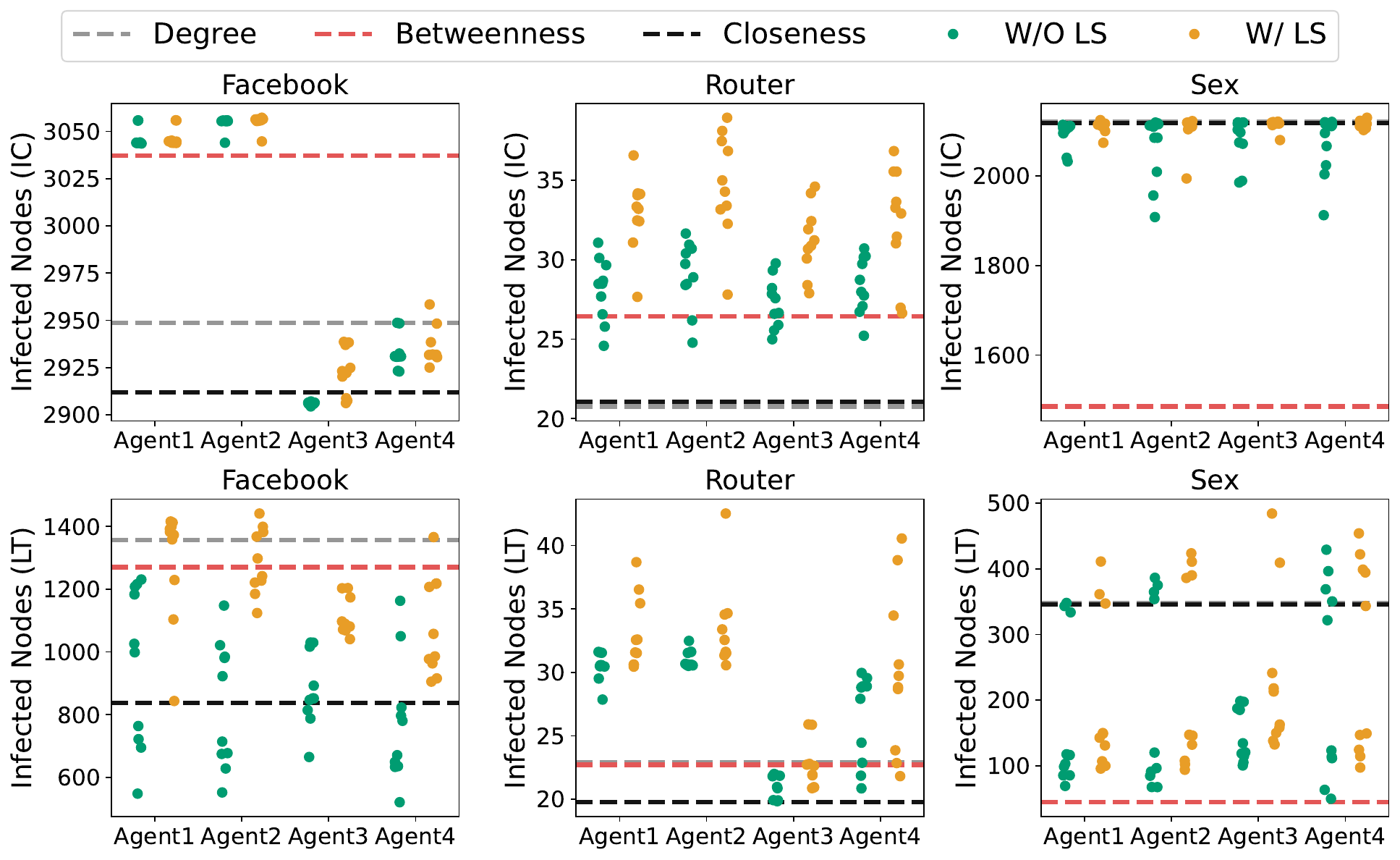}
\caption{The IM result of MLLM agents with and without local search on large-scale networks.}
\label{distribution_large}
\end{figure}

\textbf{Mixed agents with different strategies can be easily adapted to various scenarios:} The variation in performance across different networks, as seen with Agent 3 being the worst performer in the Router network but the best in the Sex network, suggests that different agents are better suited for specific types of network topologies. This observation implies that no single strategy is universally optimal across all scenarios. A combination of agents with different selection biases could provide a more robust and adaptable approach, leveraging the strengths of each agent based on the network's unique structure. It is to be expected that more sophisticated agents will achieve better performance in the future.

\begin{figure*}
        \centering
	\ContinuedFloat*
	\includegraphics[height=4cm,width=16cm]{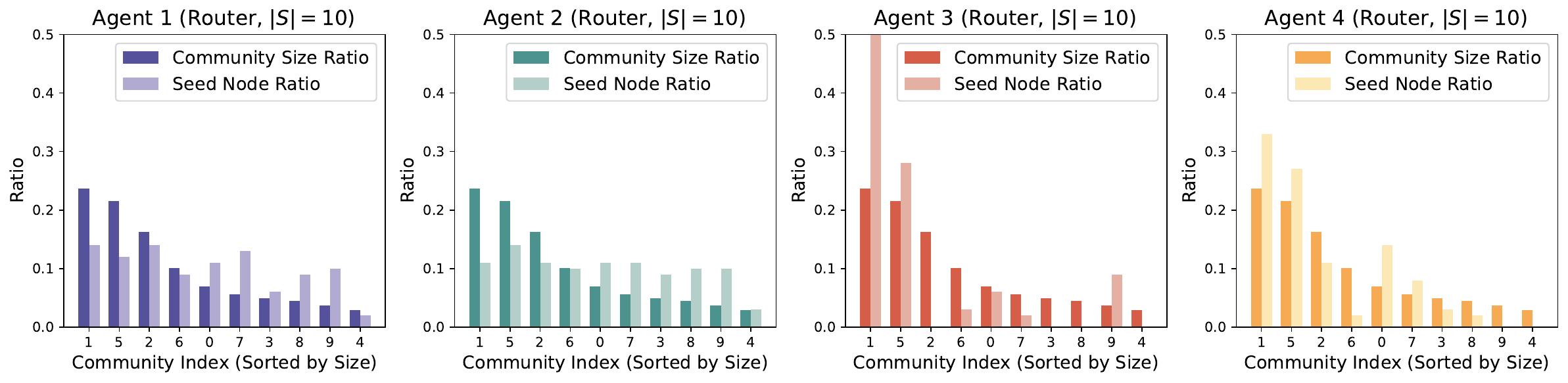}
	\caption{The distribution of selected seed nodes by different agents on the Router network. The bars refer to the community size ratio (darker bars) alongside the seed node ratio (lighter bars) for community indices sorted by size.}
    \label{sample_router}
\end{figure*}

\begin{figure*}
    \centering
	\ContinuedFloat
	\includegraphics[height=4cm,width=16cm]{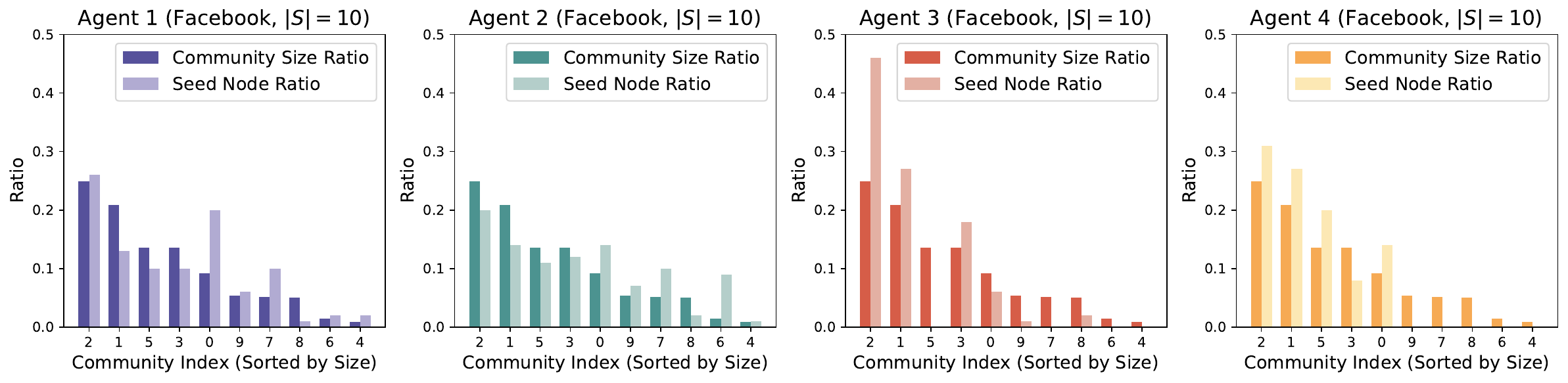}
	\caption{The distribution of selected seed nodes by different agents on the Facebook network. The bars refer to the community size ratio (darker bars) alongside the seed node ratio (lighter bars) for community indices sorted by size.}
    \label{sample_facebook}
\end{figure*}

\begin{figure*}
    \centering
	\ContinuedFloat
	\includegraphics[height=4cm,width=16cm]{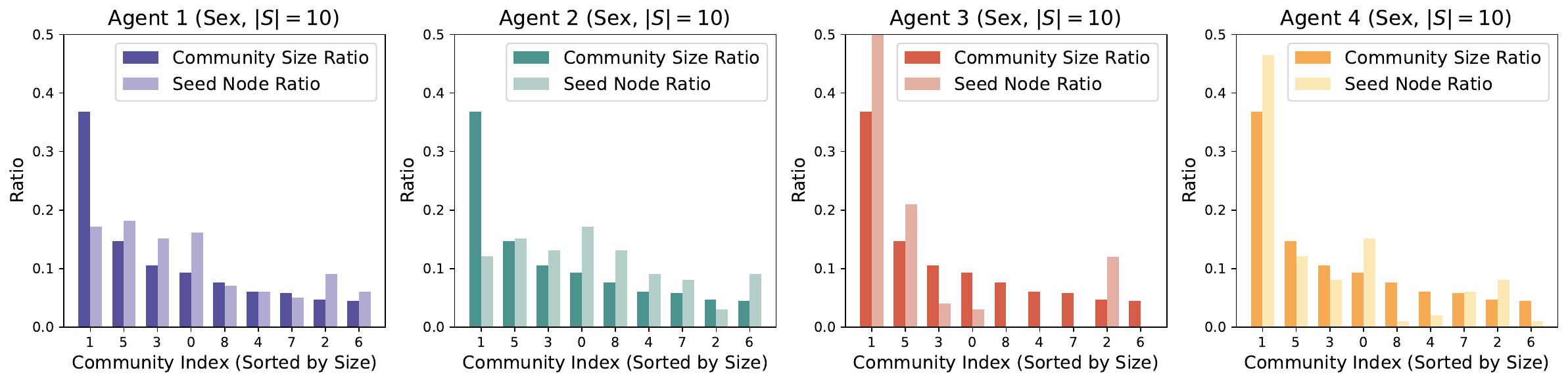}
	\caption{The distribution of selected seed nodes by different agents on the Sex network. The bars refer to the community size ratio (darker bars) alongside the seed node ratio (lighter bars) for community indices sorted by size.}
\label{sample_sex}
\end{figure*}

\textbf{MLLM exhibits an excellent spatial awareness:} Figure \ref{sample_router}(a) presents the distribution of sampled nodes by four different MLLM agents in the Router network, each with a seed size of 10.  MLLM exhibits spatial intelligence, as seen in Agent 1, which operates without specific guidance yet still distributes seed nodes in a balanced manner across communities. Furthermore, the results show that the MLLM agents can accurately follow the specific guidance provided to them. For example, Agent 2, tasked with distributing nodes proportionally across communities, adheres closely to the community size ratio. The results also reveal that certain agents, such as Agents 3 and 4, rarely select any seed nodes from certain communities. This is particularly evident in smaller communities where these agents' biases led them to focus primarily on larger or more central communities. Agent 3, with its emphasis on central nodes in the image, and Agent 4, which prioritizes large communities, both completely overlooked some of the smaller communities in the network. Figures \ref{sample_facebook}(b) and \ref{sample_sex}(c) present the results of the distribution of selected seeds by different agents on Facebook and Sex networks, showing similarities to Router's results.

\begin{figure*}[htbp]
\centering
\includegraphics[height=4.8cm,width=17cm]{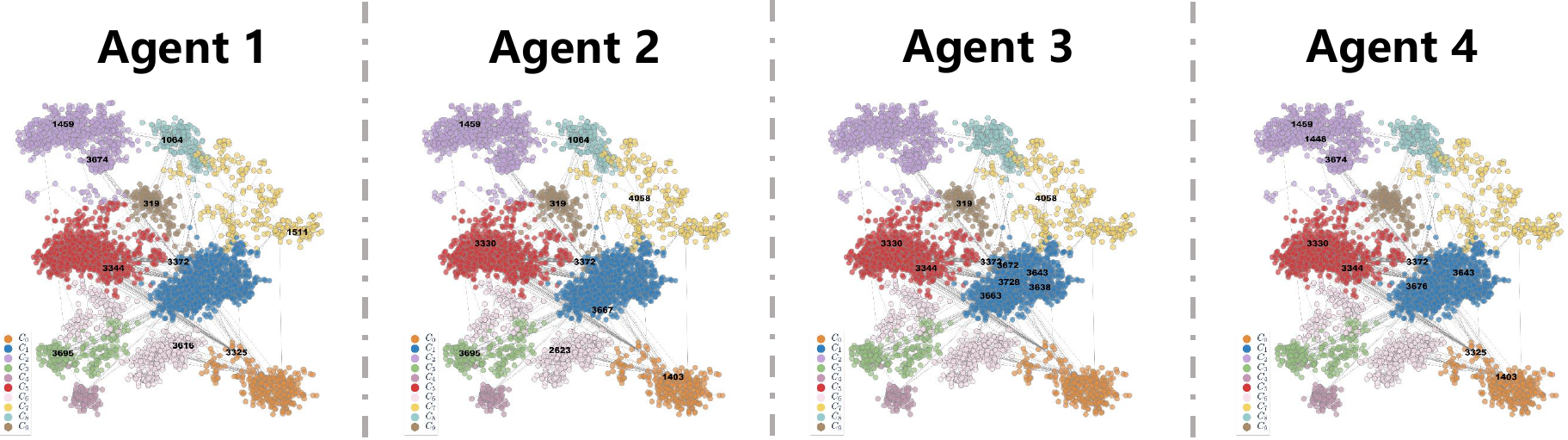}
\caption{The seeds selected by different agents on Router. Agent 1: Intelligent Selector; Agent 2: Community-Aware Selector; Agent 3: Center-Place Selector; Agent 4: Large Community Selector.} 
\label{selection_router_10}
\end{figure*}

\textbf{MLLM possesses a deep understanding of graph problems without any fine-tuning:} Some selection seeds of the different agents are shown in Figure \ref{selection_router_10}. Agent 1 takes into account both the diversity of the selection area and the avoidance of selecting nodes from small and peripheral communities (as can be seen from the low seed node ratios in communities 4 in Figure \ref{sample_router}). These aspects are exactly the core idea of Agents 2, 3 and 4, which are guided by humans.

\section{Network dismantling}\label{sec.dismantle}

\textbf{Network Dismantling (ND)} ais to identify a minimal set of nodes \( S \subset V \) whose removal causes a significant reduction in the size of the largest connected component, effectively fragmenting the network. Given a network with \( N \) nodes, the robustness  defined as: 
$
R = \frac{1}{N} \sum_{Q=1}^{N} s(Q),
$
where \( s(Q) \) represents the size of the largest connected component after the removal of \( Q \) nodes.

\subsection{Benchmark}
In the comparative study of network dismantling, two commonly used benchmarks are included:

\textbf{High-degree (HD)}  repeatedly identifying and removing the node with the highest degree in the remaining network. This process is dynamic, as the degree of nodes changes after each removal, ensuring that the most connected node at each step is eliminated. 

\textbf{High-collective influence (HCI)} is similar to HD, where at each step, the node with the highest collective influence in the remaining network is removed.

\subsection{Experimental result}
\begin{table*}[h!]
\centering
\caption{Context-setting and output directive prompts for network dismantling. Context-setting is placed at the beginning of the prompt to explain the input information and the played role to agents and the output is placed at the end of the prompt to restrict the output format.}
\label{table:prompts_GD}
\begin{tabular}{m{3.5cm}m{5.5cm}m{5.5cm}}
\Xhline{5\arrayrulewidth}
\textbf{Task} & \textbf{Context-setting prompt} & \textbf{Output directive prompt} \\ \hline
\textbf{Graph dismantling} & You are an expert in network science and you will be provided with a network in the form of an image. Each node is labeled with its node id in black text. & Do NOT output any other text or explanation. Just tell me the node id only. Your answer should be: node id. \\ \hline
\Xhline{5\arrayrulewidth}
\end{tabular}
\end{table*}

\begin{figure*}[htbp]
\centering
\includegraphics[height=10.8cm,width=16.9cm]{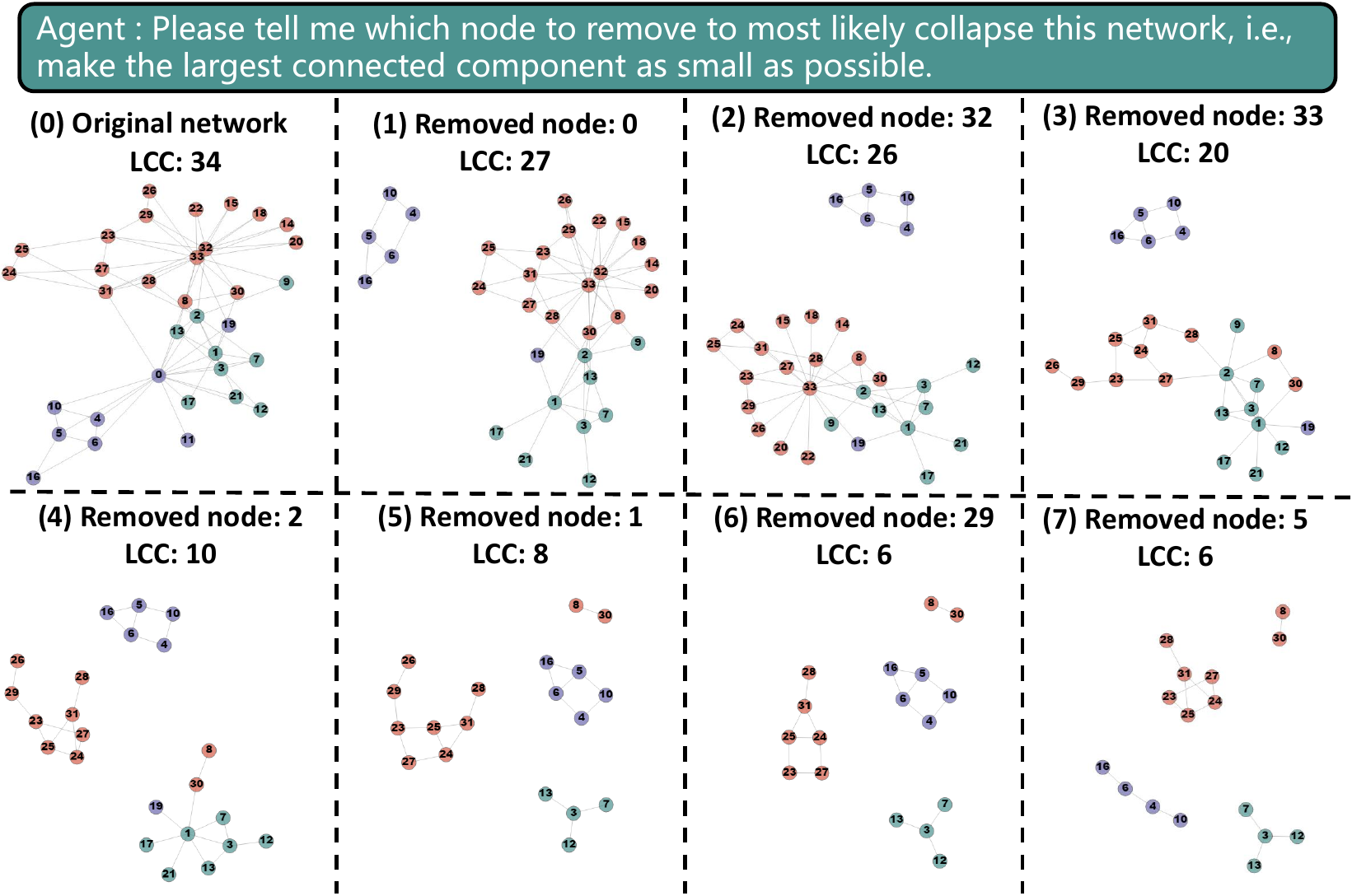}
\caption{The diagram of network dismantling guided by MLLM on the Karate network. The network is iteratively fed into the MLLM as an image to obtain suggestions for the next node to remove. The layout will dynamically adjust in response to changes in the network structure.} 
\label{GD-diagram}
\end{figure*}

\textbf{MLLMs possess a strong grasp of graph structure:} The prompt for network dismantling can be found in Table \ref{table:prompts_GD} and Figure \ref{GD-diagram} where the latter also illustrates an attempt of the network dismantling process guided by an MLLM. In traditional approaches like degree centrality, the nodes with the highest degree, such as 32 or 33, would be prioritized for removal to minimize the size of the largest connected component (LCC). However, the MLLM suggests removing node 0 first, which leads to a more rapid reduction in the LCC size, immediately to 27. This result implies the MLLM's ability to predict the cascading effects of node removal beyond the most intuitive observation (degree).

\begin{figure*}[htbp]
\centering
\includegraphics[height=3.9cm,width=17cm]{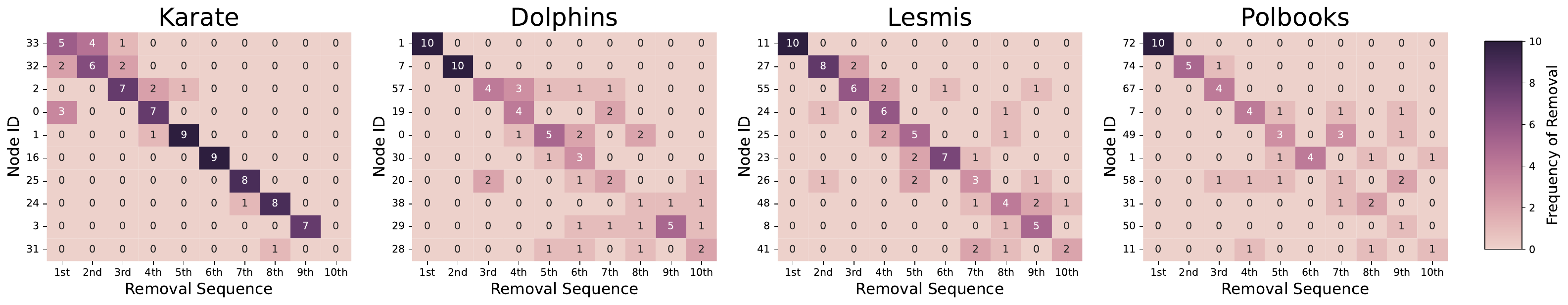}
\caption{The frequency of node removal using MLLMs for network dismantling. Each cell shows the frequency with which each node (y-axis) was removed at a particular sequence position (x-axis) over ten tests.} 
\label{Dismantle_sequence}
\end{figure*}

\textbf{Network size will affect the decision robustness of MLLMs:} In the Karate network, the MLLMs show a relatively concentrated pattern of node removal, reflected by the dark color of the diagonal elements in Figure \ref{Dismantle_sequence}. The growing size and complexity of networks likely hinder the MLLMs' ability to pinpoint a single set of critical nodes such as Polbooks. The differing removal frequencies suggest that the MLLMs' selections will be more varied, likely due to the difficulties in visual identification.

\begin{figure*}[htbp]
\centering
\includegraphics[height=4.8cm,width=17cm]{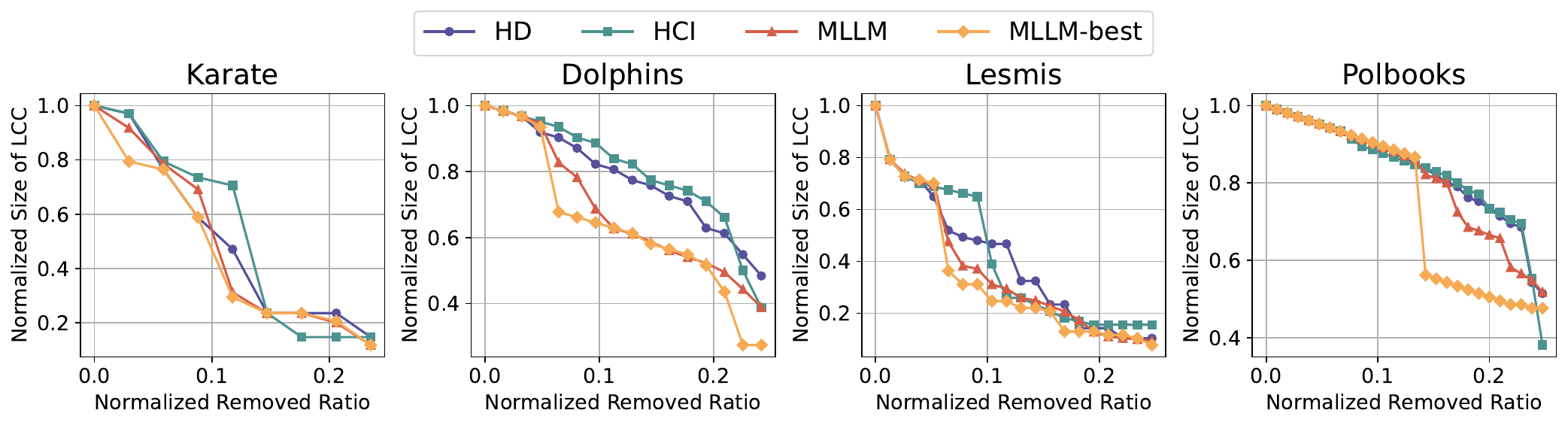}
\caption{The comparative performance on the normalized size of the Largest Connected Component (LCC) of four methods in network dismantling. MLLM refers to the average performance over ten attempts using Multi-Modal Large Language Models and MLLM-best is the best result among ten attempts. The dismantling process stops after 25\% nodes are removed.} 
\label{dismantle}
\end{figure*}

\textbf{MLLMs can beat traditional methods with its inherent intelligence:} Figure \ref{dismantle} presents the results of network dismantling on different networks. Note that the MLLMs are currently only applicable to the full-label case due to the lack of interactive channels between MLLMs and the visualization tools. The results demonstrate that both the MLLM and MLLM-best consistently outperform traditional methods such as HD and HCI in reducing the LCC size.

\begin{table*}[]
\caption{The area under the curve (AUC) of different node removal strategies across networks.}
\label{tab.dismantle}
\centering
\begin{tabular}{lccccccc}
\Xhline{5\arrayrulewidth} 
\textbf{Network} & \textbf{Karate} & \textbf{Dolphins} & \textbf{Lesmis} & \textbf{Polbooks} \\
\hline
Degree & 4.07    & 11.77    & 7.62   & 21.85  \\
CI & 4.31    & 12.13   & 7.80  & 21.81  \\
MLLM & 3.94    & 10.28   & 6.88  & 21.27  \\
MLLM-best & \textbf{3.67}    & \textbf{9.67}   & \textbf{6.33}  & \textbf{19.41}  \\
\Xhline{5\arrayrulewidth}
\end{tabular}
\end{table*}

Table \ref{tab.dismantle} presents the AUC for the normalized size of the LCC (in Figure \ref{dismantle}) with lower AUC values indicating better result. Not only the MLLM-best but also MLLM consistently shows the lowest AUC across networks, demonstrating its effectiveness in network dismantling.

\section{MLLM on basic graph-related tasks}\label{sec.basic}
In this section, we will investigate the MLLM on some basic graph-structured tasks and identify factors affecting the performance of MLLM.

\begin{table*}[]
\captionof{table}{Structural Metrics of Synthetic Networks under two configurations: Including Average Node and Edge Counts, Degree, Shortest Distance, Connected Components, and Cycle Presence Proportion. }
\label{tab.random}
\centering
\begin{tabular}{lcccccc}
\Xhline{5\arrayrulewidth}
\multirow{2}{*}{\textbf{Metrics}} & \multicolumn{2}{c}{\textbf{WS}} & \multicolumn{2}{c}{\textbf{BA}} & \multicolumn{2}{c}{\textbf{ER}} \\
\cmidrule(lr){2-3} \cmidrule(lr){4-5} \cmidrule(lr){6-7}
& \textbf{Easy}    & \textbf{Hard}    & \textbf{Easy}    & \textbf{Hard}    & \textbf{Easy}    & \textbf{Hard}    \\
\midrule
\textbf{Avg. Nodes} & 7.58 & 17.58 & 7.69  & 17.51& 12.63& 17.39\\
\textbf{Avg. Edges}& 7.58& 17.58& 12.38 & 32.01  & 15.01& 14.10 \\
\textbf{Avg. Degree} & 2.00& 2.00  & 3.18 & 3.65& 2.33 &1.61 \\
\textbf{Avg. Shortest Dist.}& 1.61& 1.57& 1.547& 2.08& 0.82  & 0.11\\
\textbf{Avg. Component} & 1.27& 1.83& 1.00& 1.00& 2.20       & 5.15\\
\textbf{Cycle Existence} & 100.0\% & 100.0\%& 100.0\%        & 100.0\% & 100.0\% & 100.0\%\\
\Xhline{5\arrayrulewidth}
\end{tabular}
\end{table*}

\subsection{Synthetic network}\label{sec.synthetic}

Three types of random networks are utilized: Barabási-Albert (BA) network, Erdős-Rényi (ER) network and Watts-Strogatz (WS) network. Table \ref{tab.random} lists the structural information of these networks where BA is viewed as dense network and WS and ER are relatively sparse sometimes containing multiple connected components. 

\begin{itemize}
\item \textbf{Erdős-Rényi (ER)} network model is a foundational concept in random graph theory \citep{erdos1960evolution}. In an ER network, a graph is constructed by connecting nodes randomly with a given probability $p$. 

\item \textbf{Barabási-Albert (BA)} model generates scale-free networks featured by a power-law degree distribution \citep{barabasi1999emergence}. 

\item \textbf{Watts-Strogatz (WS)} network exhibits high clustering and short average path lengths \citep{watts1998collective}. The WS model starts with a regular ring lattice where each node is connected to $k$ nearest neighbors and with a probability $p$, each edge is randomly rewired then. 
\end{itemize}

The visualization of networks has different layouts. In this work, we have tested three types to investigate the influence of layouts on the effectiveness of MLLMs.

\begin{itemize}
\item \textbf{Fruchterman-Reingold Layout} is a force-directed algorithm that simulates physical forces between the nodes and edges of a graph. Nodes repel each other like charged particles, while edges act like springs that pull connected nodes together, to minimize edge crossings and evenly distribute them.

\item \textbf{Circle Layout} denotes the layout that all nodes are placed at equal distances from each other along the circumference of a circle. 

\item \textbf{Grid Layout} arranges nodes in a regular grid pattern, with each node occupying a unique position. This layout is effective for displaying nodes in a structured, non-overlapping manner, making it easier to compare their positions and relationships. 
\end{itemize}

\begin{figure*}[htbp]
\centering
\includegraphics[height=3.4cm,width=17cm]{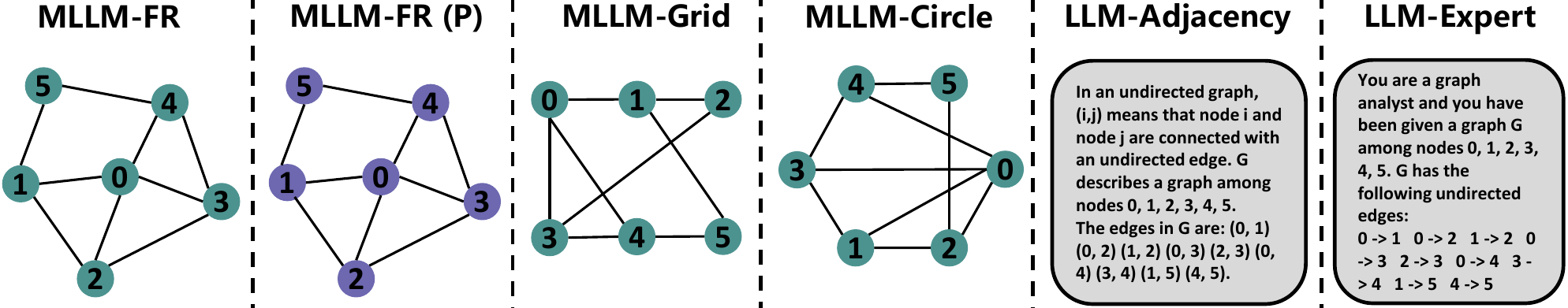}
\caption{Illustrations of various representations of the same graph with six nodes, including input images with different layouts and colors provided to the MLLM, accompanied by the sentence, ``You are an expert in network science and will be provided with a network G in the form of an image," along with two types of textual descriptions.} 
\label{MLLM_ability}
\end{figure*}

To further assess the understanding capabilities of MLLMs on graph structures, we evaluate them on six fundamental graph problems. In addition to the image and leading sentence (see Figure \ref{MLLM_ability}), the problem itself is also included as part of the prompt, as detailed in Table \ref{table:question}.

\begin{table*}[]
\centering
\caption{The prompt used to evaluate MLLMs' capabilities on six fundamental graph-structured problems.}
\label{table:question}
\begin{tabular}{m{4cm}m{11.5cm}}
\Xhline{5\arrayrulewidth}
\textbf{Problem} & \textbf{Question}  \\ \hline
\textbf{Node Degree} & Given the network G provided, please answer the following question: How many connections does node 1 have? The answer is a number, denoted as A1. Your output should be a list as [A1] without any text and explanation.  \\ \hline
\textbf{Highest Degree Node} & Given the network G provided, please answer the following question: Which node has the highest degree value? The answer is a number, denoted as A1. Your output should be a list as [A1] without any text and explanation.  \\ \hline
\textbf{Highest Betweenness Node} & Given the network G provided, please answer the following question: Which node has the highest betweenness value? The answer is a number, denoted as A1. Your output should be a list as [A1] without any text and explanation.  \\ \hline
\textbf{Shortest Distance} & Given the network G provided, please answer the following question: What is the shortest distance between node 1 and node 2? The answer is a number or False if they cannot reach each other, denoted as A1. Your output should be a list as [A1] without any text and explanation.  \\ \hline
\textbf{Cycle Detection} & Given the network G provided, please answer the following question: Does the network contain a cycle? The answer is either True or False, denoted as A1. Your output should be a list as [A1] without any text and explanation.  \\ \hline
\textbf{Connected Components} & Given the network G provided, please answer the following question: How many connected components does the network have? The answer is a number, denoted as A1. Your output should be a list as [A1] without any text and explanation.  \\ 
\Xhline{5\arrayrulewidth}
\end{tabular}
\end{table*}

\begin{table*}[ht]
\centering
\caption{The capability of different models on the basic graph-structured task. 
Task 1 (Node Degree): Calculate the degree of a specific node;
with the highest betweenness centrality;
Task 2 (Highest Degree Node): Identify the node with the highest number of connections;
Task 3 (Highest Betweenness Node): Identify the node 
Task 4 (Shortest Distance): Determine the shortest path between two specified nodes;
Task 5 (Cycle Detection): Identify whether the network contains a cycle.
Task 6 (Connected Components): Identify the number of distinct connected components.}
\begin{subtable}{\textwidth}
    \centering
    \resizebox{\textwidth}{!}{
    \begin{tabular}{lcccccccccccccc}  
    \Xhline{5\arrayrulewidth}  
    \multirow{2}{*}{\textbf{Model}}&  
    \multicolumn{2}{c}{\textbf{Task 1}}&\multicolumn{2}{c}{\textbf{Task 2}}&\multicolumn{2}{c}{\textbf{Task 3}}&\multicolumn{2}{c}{\textbf{Task 4}}&\multicolumn{2}{c}{\textbf{Task 5}}&\multicolumn{2}{c}{\textbf{Task 6}}\cr  
    \cmidrule(lr){2-3} \cmidrule(lr){4-5} \cmidrule(lr){6-7}\cmidrule(lr){8-9}  \cmidrule(lr){10-11} \cmidrule(lr){12-13}
    & \textbf{Easy}& \textbf{Hard}& \textbf{Easy}& \textbf{Hard}& \textbf{Easy}& \textbf{Hard}& \textbf{Easy}& \textbf{Hard}& \textbf{Easy}& \textbf{Hard}& \textbf{Easy}& \textbf{Hard}\cr 
    \midrule   
    LLM-Expert & 89.5\% & 74.0\% & {98.5\%} & \textbf{92.5\%} & 72.0\% & {72.5\%} & \textbf{89.5\%} & \textbf{58.0\%} & \textbf{100.0\%} & \textbf{100.0\%} & \textbf{100.0\%} & \textbf{100.0\%}\\
    LLM-Adjacency & \textbf{96.0\%} & \textbf{76.5\%} & \textbf{99.5}\% & 91.0\% & 71.0\% & \textbf{75.5\%} & {86.0\%} & 51.0\% & \textbf{100.0\%} & \textbf{100.0\%} & \textbf{100.0\%} & 99.5\%\\
    MLLM-FR & 54.5\% & 36.0\% & 88.5\% & 77.5\% & {77.5\%} & 69.0\% & 62.5\% & 39.0\% & \textbf{100.0\%} & \textbf{100.0\%} & \textbf{100.0\%} & \textbf{100.0\%}\\
    MLLM-FR(P) & 63.0\% & 42.5\% & 88.5\% & 79.5\% & \textbf{80.0\%} & 66.5\% & 60.5\% & 40.5\% & \textbf{100.0\%} & \textbf{100.0\%} & \textbf{100.0\%} & \textbf{100.0\%}\\
    MLLM-Circle & 19.0\% & 11.5\% & 91.5\% & 59.0\% & 75.5\% & 53.5\% & 63.0\% & {45.0\%} & 99.5\% & \textbf{100.0\%} & \textbf{100.0\%} & \textbf{100.0\%}\\
    MLLM-Grid & 26.0\% & 10.0\% & 64.0\% & 25.5\% & 43.0\% & 16.5\% & 53.0\% & {48.5\%} & \textbf{100.0\%} & \textbf{100.0\%} & 98.5\% & 99.5\%\\
    \Xhline{5\arrayrulewidth}   
    \end{tabular}
    }
    \caption{Barabási-Albert (BA) network. The number of edges each new node connects to when it is added to the network is set to 2. \#Easy: $n \in [5, 10]$; \#Hard: $n \in [15, 20]$.}
\end{subtable}

\vspace{0.3em} % Adds some vertical space between subtables

\begin{subtable}{\textwidth}
    \centering
    \resizebox{\textwidth}{!}{
    \begin{tabular}{lcccccccccccccc}  
    \Xhline{5\arrayrulewidth}  
    \multirow{2}{*}{\textbf{Model}}&  
    \multicolumn{2}{c}{\textbf{Task 1}}&\multicolumn{2}{c}{\textbf{Task 2}}&\multicolumn{2}{c}{\textbf{Task 3}}&\multicolumn{2}{c}{\textbf{Task 4}}&\multicolumn{2}{c}{\textbf{Task 5}}&\multicolumn{2}{c}{\textbf{Task 6}}\cr  
    \cmidrule(lr){2-3} \cmidrule(lr){4-5} \cmidrule(lr){6-7}\cmidrule(lr){8-9}  \cmidrule(lr){10-11} \cmidrule(lr){12-13}
    & \textbf{Easy}& \textbf{Hard}& \textbf{Easy}& \textbf{Hard}& \textbf{Easy}& \textbf{Hard}& \textbf{Easy}& \textbf{Hard}& \textbf{Easy}& \textbf{Hard}& \textbf{Easy}& \textbf{Hard}\cr 
    \midrule  
LLM-Expert & 88.0\% & {94.0\%} & \textbf{91.5\%} & 90.0\% & 59.5\% & 64.0\% &{64.0\%} & {66.5\%} & 74.0\% & 49.0\% & 23.5\% & 26.0\%\\
LLM-Adjacency & \textbf{95.5\%} & \textbf{94.5\%} & 91.0\% & \textbf{94.5\%} & 62.5\% & {70.0\%} & 60.5\% & 60.5\% & 93.0\% & 82.5\% & 32.0\% & 26.0\%\\
MLLM-FR & 76.0\% & 81.5\% & 81.5\% & 84.0\% & \textbf{68.0\%} & 67.0\% & \textbf{65.0\%}& \textbf{67.0}\% & 86.0\% & 75.5\% & \textbf{93.0\%} & \textbf{54.5\%}\\
MLLM-FR(P) & 73.0\% & 82.0\% & 80.0\% & 91.0\% & 65.0\% & \textbf{77.0\%} & 52.0\% & 61.5\% & 89.0\% & 72.0\% & 87.0\% & \textbf{54.5\%}\\
MLLM-Circle & 21.5\% & 12.5\% & {73.0\%} & 72.0\% & 44.0\% & 45.5\% & 32.0\% & 15.5\% & {98.5\%} & \textbf{97.0\%} & 43.5\% & 5.0\%\\
MLLM-Grid & 19.5\% & 17.5\% & 49.0\% & 47.0\% & 20.0\% & 24.5\% & 34.5\% & 22.5\% & \textbf{99.0\%} & \textbf{97.0\%} & 45.5\% & 4.5\%\\
    \Xhline{5\arrayrulewidth}   
    \end{tabular}
    }
    \caption{Erdős-Rényi (ER) network. The probability that any pair of nodes will have an edge connecting them is set to 0.2 for the easy case and 0.1 for the hard case. \#Easy: $n \in [10, 15]$; \#Hard: $n \in [15, 20]$.}
\end{subtable}

\vspace{0.3em} % Adds some vertical space between subtables

\begin{subtable}{\textwidth}
    \centering
    \resizebox{\textwidth}{!}{
    \begin{tabular}{lcccccccccccccc}  
    \Xhline{5\arrayrulewidth}  
    \multirow{2}{*}{\textbf{Model}}&  
    \multicolumn{2}{c}{\textbf{Task 1}}&\multicolumn{2}{c}{\textbf{Task 2}}&\multicolumn{2}{c}{\textbf{Task 3}}&\multicolumn{2}{c}{\textbf{Task 4}}&\multicolumn{2}{c}{\textbf{Task 5}}&\multicolumn{2}{c}{\textbf{Task 6}}\cr  
    \cmidrule(lr){2-3} \cmidrule(lr){4-5} \cmidrule(lr){6-7}\cmidrule(lr){8-9}  \cmidrule(lr){10-11} \cmidrule(lr){12-13}
    & \textbf{Easy}& \textbf{Hard}& \textbf{Easy}& \textbf{Hard}& \textbf{Easy}& \textbf{Hard}& \textbf{Easy}& \textbf{Hard}& \textbf{Easy}& \textbf{Hard}& \textbf{Easy}& \textbf{Hard}\cr 
    \midrule  
    LLM-Expert & \textbf{98.5\%} & 94.5\% & \textbf{99.0\%} & 92.5\% & 78.0\% & 43.0\% & 76.5\% & 47.0\% & 84.0\% & 92.5\% & 57.5\% & 29.5\%\\
    LLM-Adjacency & 95.5\% & \textbf{95.5\%} & \textbf{99.0\%} & \textbf{98.5\%} & 73.0\% & 53.5\% & \textbf{80.0\%} & 36.0\% & \textbf{100.0\%} & \textbf{100.0\%} & 70.5\% & 33.0\%\\
    MLLM-FR & 81.5\% & 66.5\% & 96.5\% & 82.0\% & \textbf{90.0\%} & 57.0\% & 69.0\% & {47.0\%} & 91.0\% & 90.0\% & 99.0\% & 89.5\%\\
    MLLM-FR(P) & 77.5\% & 74.0\% & 97.5\% & 88.5\% & 88.5\% & \textbf{68.0\%} & 58.5\% & \textbf{50.0\%} & 89.5\% & 85.0\% & \textbf{100.0\%} & \textbf{93.5\%}\\
    MLLM-Circle & 64.5\% & 47.5\% & 90.5\% & 70.5\% & 70.5\% & 32.0\% & 52.5\% & 26.0\% & 98.0\% & \textbf{100.0\%} & 93.0\% & 43.0\%\\
    MLLM-Grid & 27.5\% & 21.0\% & 63.5\% & 50.0\% & 43.5\% & 20.0\% & 49.5\% & 23.0\% & 97.0\% & 98.5\% & 85.5\% & 50.5\%\\
    \Xhline{5\arrayrulewidth}   
    \end{tabular}
    }
    \caption{Watts-Strogatz (WS) network. The number of nearest neighbors each node is connected to in the initial ring lattice is set to 1 and the probability of rewiring each edge is set to 0.2. \#Easy: $n \in [5, 10]$; \#Hard: $n \in [15, 20]$.}
\end{subtable}
\end{table*}

\textbf{MLLM excels in tasks requiring global awareness:} The performance of MLLM-FR and MLLM-FR(P) in tasks 3 (Highest Betweenness Node), 4 (Shortest Distance), and 6 (Connected Components) showcases their ability to handle problems that require a comprehensive understanding of the entire network structure. MLLM's ability to process these global relationships efficiently leads to its dominance over other methods in such tasks.

\textbf{The color has minimal impact on MLLM's performance:} The close similarity in results between MLLM-FR and MLLM-FR(P) demonstrates that the color of visual representations has little influence on the model's effectiveness since both layouts provide nearly identical performance across the tasks.

% it suggests that MLLM focuses more on the structure and spatial relationships of nodes rather than being swayed by color variations in the input images.

\textbf{Layout greatly affects performance:} The difference between the results of MLLM-FR and models using MLLM-Circle or MLLM-Grid layouts highlights the importance of the layout. MLLM-FR, which uses a force-directed layout, provides a clearer visual network structure, leading to superior performance. In contrast, MLLM-Circle and MLLM-Grid offer less intuitive spatial arrangements, making it harder for the model to recognize global features, leading to poorer results across tasks. Moreover, some layouts even lost some basic structural information, for example, the connection of node 0 and node 2 cannot reflected in the grid case of Figure \ref{MLLM_ability}.

\textbf{MLLM's adaptability across different network structures:} MLLM maintains performance in global tasks (3 and 6) regardless of network density, as evidenced by its comparable results in both sparse networks like ER and WS and denser networks like BA. In contrast, LLM shows a marked drop in performance, particularly in sparser networks, where spatial awareness is crucial for success. MLLM’s ability to retain its effectiveness across these varying structures highlights its suitability for tasks that require a broader perspective, where LLM struggles due to its localized understanding.

\textbf{MLLM's strength over LLM in large-scale problems:} The superior performance of MLLM in tasks requiring global awareness suggests that it is better equipped to handle large-scale problems where a comprehensive understanding of the entire network is essential. Furthermore, LLM’s reliance on extensive natural language prompts when encoding large-scale graphs further limits its capability, making MLLM a more suitable choice for tasks that involve larger, more complex network structures.

\section{Discussion and Prospect}\label{sec.discussion}
In addition to the aforementioned spatial intelligence of MLLMs on graph-structured problems, another key strength of MLLMs lies in their remarkable scalability, which is particularly advantageous when dealing with large-scale networks. Real-world networks are typically massive \citep{leskovec2016snap}, making it impractical to encode the entire network into a text-based prompt. In contrast, by leveraging visual inputs in the form of network images, MLLMs bypass this limitation. Regardless of how large or complex the network is, the input remains a fixed-size image, allowing the MLLM to interpret and process it efficiently. Unlike adjacency matrices and learned embeddings, which trade off structural information for computation, images serve as the most intuitive representation of graph structures, effectively preserving valuable high-order information such as community structures, paths, and motifs, and so on.

\begin{figure}
\centering
\includegraphics[width=0.35\textwidth]{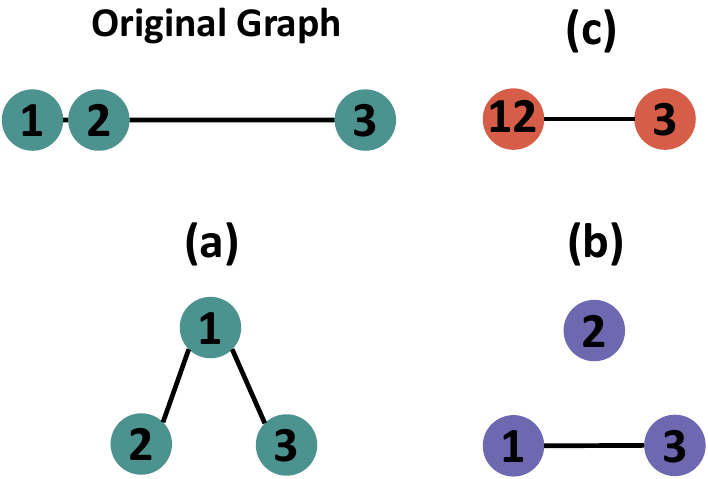}
\caption{An example of possible outcomes from MLLM recognition on the same graph.}
\label{recognition}
\end{figure}

The current MLLMs may sometimes return undesirable outcomes. Figure \ref{recognition} shows several possible recognition results of MLLMs on one graph. The original graph consists of three nodes (1, 2, and 3) where node 1 is connected to node 2, and node 2 is connected to node 3. Case (a): This is the correct recognition of the graph by the MLLMs. Case (b): The MLLMs incorrectly recognize the structure by displaying node 2 between node 1 and node 3 but fail to recognize the edge between nodes 1 and 2. Case (c): In this scenario, nodes 1 and 2 are so close to each other that the MLLMs misrecognize them as a single node labeled `12'.

As observed, MLLMs' full potential is still constrained by the lack of effective visualization tools. This is the reason why we call this representation of graph as low-loss. Even humans face difficulties in recognizing and interpreting individual nodes when a large number of them are plotted on a fixed-size canvas. In such cases, a tool analogous to a magnifying glass would allow for a more detailed, micro-level examination of specific areas of the network. This limitation in visualization should not be considered a flaw in MLLMs itself, as it reflects a broader challenge in rendering and interpreting complex, dense networks visually.

If visualization software can be seamlessly integrated with MLLMs to support interactive exploration—enabling zooming and detailed node examination in real-time, the performance and applicability of MLLMs would be greatly enhanced. This would not only improve MLLMs' reasoning capabilities on large-scale networks but also enable full-scale labeling and analysis, similar to what is currently achievable with small-scale networks. Achieving this would allow for loss-free representation of graph-structured data through images, opening a new paradigm for graph-related computations. Note that the proposed MLLM-based method is generalizable and could extend beyond the problems studied here to other challenges, such as graph coloring, vertex cover, and graph partitioning, with our present work providing a strong foundation for these future developments.

\section{Conclusion}\label{sec.conslution}
In this work, we have demonstrated the effectiveness of MLLMs in addressing complex graph-structured combinatorial problems, such as network dismantling and influence maximization. By utilizing simple prompts combined with local search strategies, our approach achieves superior performance over traditional methods and GNN-based approaches. We provided a comprehensive analysis of MLLMs' capabilities on fundamental graph tasks and identified key factors that enhance their effectiveness. Our findings reveal the potential of MLLMs to revolutionize large-scale graph problem-solving, marking a significant step toward harnessing their full capacity in practical, real-world applications. Our Future work will explore integrating visualization tools with MLLMs for interactive graph exploration, enhancing reasoning on large networks and enabling comprehensive analysis.

% \section*{Acknowledgments}
% This was was supported in part by......

%Bibliography
\bibliographystyle{unsrt}  
\bibliography{references}

\end{document}